\documentclass[conference]{IEEEtran}
\usepackage{times}

\usepackage[numbers,sort&compress]{natbib}
\usepackage{multicol}
\usepackage[bookmarks=true]{hyperref}

\usepackage{amsmath} 
\usepackage{amsfonts}
\usepackage{graphicx}
\usepackage{subfigure}
\usepackage{multirow}
\usepackage{caption}
\usepackage{float} 
\usepackage{subcaption}
\usepackage{stfloats}
\usepackage{pifont}
\usepackage{color}

\begin{document}

\title{AnyAmber: A Generalist for Versatile Anonymous Bearing and Range Based Position Tracking}



\author{
\authorblockN{
Si Wang\authorrefmark{1},
Yanmei Jiao\authorrefmark{2},
Yanjun Cao\authorrefmark{3},
Rong Xiong\authorrefmark{1},
Yue Wang\authorrefmark{1}\authorrefmark{4}
}

\authorblockA{
\authorrefmark{1}Institute of Cyber-Systems and Control, Zhejiang University, China
}

\authorblockA{
\authorrefmark{2}School of Information Science and Engineering, Hangzhou Normal University, China
}

\authorblockA{
\authorrefmark{3}Huzhou Institute of Zhejiang University, Huzhou, China
}

\authorblockA{
\authorrefmark{4}Corresponding author: ywang24@zju.edu.cn
}
}



%

\maketitle

\begin{abstract}
Position tracking based on bearing measurements and Ultra-wideband (UWB) ranging is widely used in robotic navigation tasks. However, due to variations in the number of robots, anchor configurations, UWB tag layouts, and the presence or absence of anonymous visual observations, existing methods typically rely on specially designed solvers or networks tailored to a particular localization problem.
In this work, we introduce AnyAmber, a generalist neural network for versatile anonymous bearing and range based position tracking. To model diverse and dynamic geometric constraints, a heterogeneous EGAT network is employed for unified localization and uncertainty estimation, cascaded with a differentiable hierarchical PGO for improved accuracy. Additionally, we incorporate a Sensor-Embedded GRU module for adaptive UWB bias correction and a temporal graph-based matching network for soft assignments between robots and anonymous bearings. 
By adopting a unified problem formulation, our model is jointly pretrained on a large-scale multi-task dataset encompassing diverse simulated and real-world environments. In the experiments, with fine-tuning on only a single trajectory in a target test scenario, the model achieves superior few-shot localization performance to existing methods.
Our code is available at: https://github.com/HiOnes/AnyAmber.
\end{abstract}

\IEEEpeerreviewmaketitle

\section{Introduction}
\begin{figure}[tp]
    \setlength{\belowcaptionskip}{-0.4cm}
    \centering
    \includegraphics[scale=0.28]{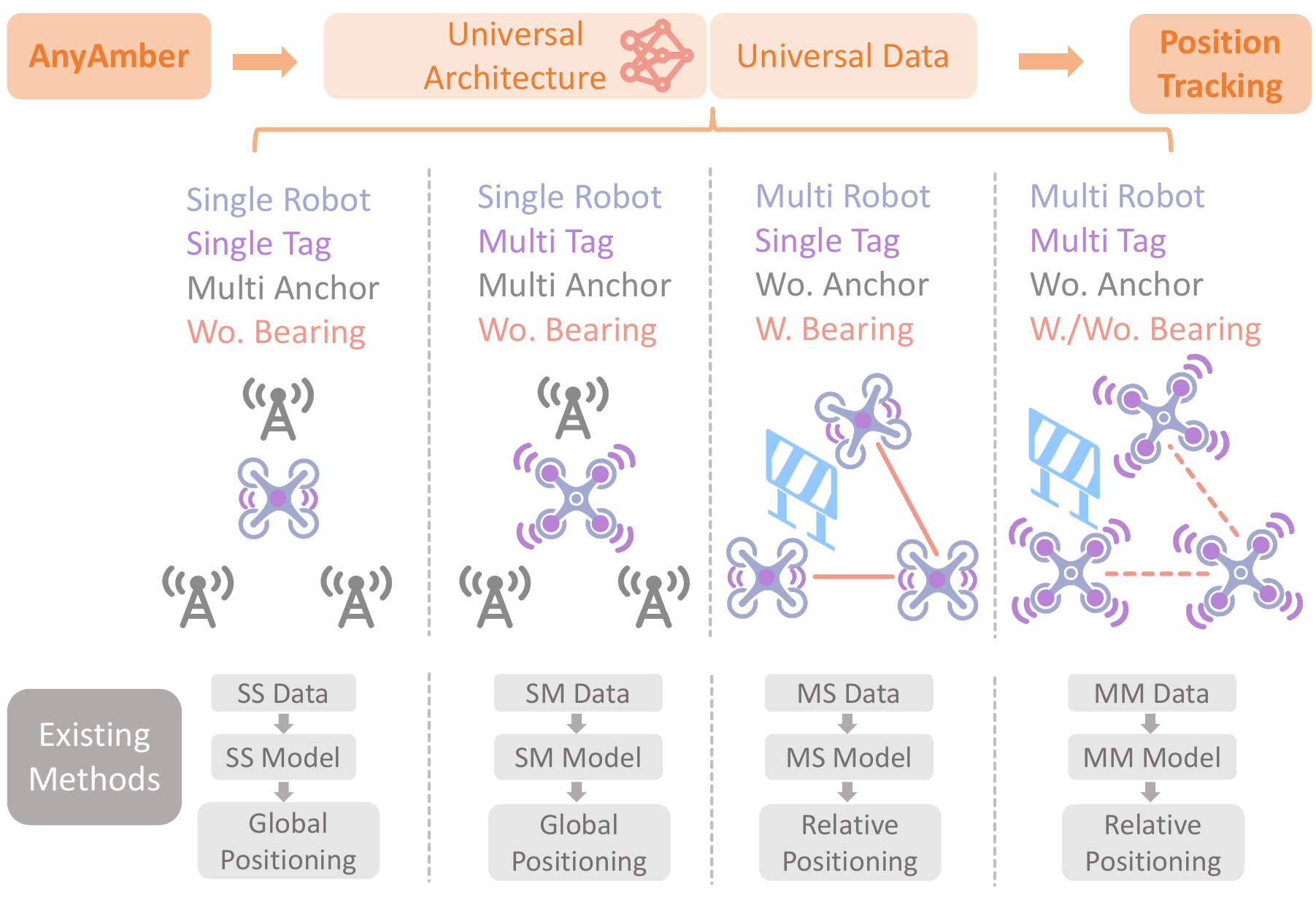}
    \caption{Overview of our universal neural localization approach.}
    \label{fig:teasing}
\end{figure}

Precise position tracking serves as a fundamental prerequisite for both single robot and robot swarms to carry out complex tasks such as exploration, search and rescue. 
In scenarios without access to global positioning aids such as GPS \cite{jaimes2008approach,qi2020cooperative,moon2016outdoor} or motion capture systems \cite {preiss2017crazyswarm}, position tracking must rely entirely on relative measurements. Bearing and range observations constitute the core modalities for establishing spatial relationships between agents and their environment.
In single-robot localization applications, range measurements between multiple fixed anchors and the robot-mounted UWB tags are utilized to estimate the robot’s position with respect to the map reference frame.
In multi-robot systems, pairwise distance measurements and anonymous visual bearings serve as constraints for estimating their relative poses in the local frame. Despite the shared geometric foundations, as shown in Fig.~\ref{fig:teasing}, these problems are often treated as distinct challenges depending on the configuration: single vs. multi-robot, or single vs. multi-tag.

\begin{figure}[tp]
    \setlength{\belowcaptionskip}{-0.2cm}
    \centering
    \includegraphics[width=.48\textwidth]{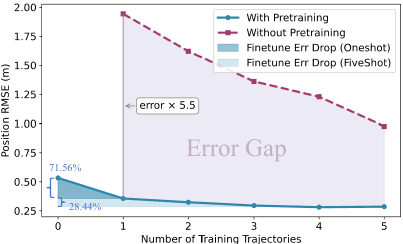}
    \caption{Localization accuracy vs. number of training trajectories for pretrained and non-pretrained models.}
    \label{fig:train-traj-num}
\end{figure}

For various tasks, existing approaches \cite{jiang2023efficient, yang2022robust, kao2023viunet, arjmandi2024deepcovpg, zheng2023enhanced, huang2024novel, he2024multi, he2025graph, wang2025neural, wang2025mr, zhao2025robust,liang20253, xun2023crepes,olson2011apriltag,yan2019active,xu2020decentralized,xu2022omni,xu2024d} have focused on optimizing performance on individual problems by designing specialized solvers or networks. Although task-specific neural networks \cite{yang2022robust, kao2023viunet, arjmandi2024deepcovpg, zheng2023enhanced, huang2024novel, he2024multi, he2025graph, wang2025neural,wang2025mr} can achieve state-of-the-art performance under particular localization settings, they lack versatility, a network designed for single-robot localization cannot easily adapt to multi-agent relative tracking. Conversely, general frameworks like Pose Graph Optimization (PGO) offer the necessary flexibility to model various configurations but often fail to match the accuracy of learning based methods when faced with non-Gaussian noise, signal multipath, or complex disturbances. Consequently, a \textit{unified} and \textit{learnable} solution capable of addressing diverse localization problems remains an open challenge.

Recent studies in visual geometry and semantics \cite{cheng2021per,cheng2022masked, jain2023oneformer, wang2025vggt} reveal that universal network architectures not only enhance model usability but also improve single-task performance through learning the generalizable features across multiple tasks. Inspired by these findings, we investigate whether the generalizable knowledge can be captured for enhanced performance through joint pretraining in the context of bearing and range based position tracking. 

To bridge the fragmentation while simultaneously boosting performance on individual tasks, in this paper, we propose \textbf{AnyAmber}, a generalist network to handle versatile anonymous bearing and range based position tracking tasks, covering single-robot single-tag (SS), single-robot multi-tag (SM), multi-robot single-tag (MS), and multi-robot multi-tag (MM) scenarios, which are illustrated in Fig. \ref{fig:teasing}. 
To achieve the above objectives, we require two key components: a unified problem formulation and high-fidelity measurements (assigned bearings, low-noise UWB rangings).
Specifically, we bridge the gap between single- and multi-robot positioning by reformulating them into a unified relative localization problem. Built upon this consistent formulation, a heterogeneous EGAT network is utilized to aggregate multi-sensor geometric features, yielding an accurate estimation of the robot’s pose and covariance. For data association between anonymous visual observations and robots, we design a temporal graph attentional neural network in conjunction with a multi-frame Sinkhorn algorithm, ensuring accurate and reliable matching. To compensate for UWB signal disturbances, a Sensor-Embedded GRU network is employed for adaptive ranging bias correction. Moreover, to improve overall accuracy while enabling end-to-end training, we incorporate a two-stage differentiable PGO scheme.

Empowered by the universal architecture, our model gains transferable localization capability through joint pretraining on a large-scale unified dataset, covering 108 diverse scenarios, 225 trajectories and 153555 sampled frames.
Comprehensive experiments verify that with only a single trajectory for fine-tuning in a specific unseen environment, the proposed AnyAmber outperforms the existing specialized models, significantly reducing the demand for data collection and boosting performance on individual tasks, as few-shot performance presented in Fig. \ref{fig:train-traj-num}. Overall, our major contributions are as follows:
\begin{itemize}
  \item Unified formulation covering various localization tasks: single- and multi-robot settings, single- and multi-tag configurations, with or without anchor setups, and bearing or non-bearing measurements.
  \item Generalist graph attention network \textbf{AnyAmber} for training and inference of versatile anonymous bearing and range based position tracking tasks.
  \item Modules to improve the performance, including GRU for UWB ranging denoising, temporal GNN-based match network for soft data association, and a hierarchical differentiable PGO structure for optimization.
  \item Substantial experimental results validating the cross-task training enables superior performance in previously unseen environments over task-specific methods, requiring only few-shot fine-tuning.
\end{itemize}

\section{Related Work}

In this section, we introduce several task-specific position tracking architectures based on visual and range measurements, as well as a set of universal multi-task models.

\subsection{Specialized Position Tracking Architecture}

\subsubsection{Single-robot Positioning}
Single-robot localization aims to recover the robot's position in a global frame from UWB tag-to-anchor ranges, possibly fused with IMU measurements. Current solutions generally fall into four classes: extended Kalman filter (EKF) frameworks \cite{li2018accurate,cao2020accurate}, nonlinear optimization formulations \cite{jiang2023efficient,hashim2023nonlinear}, B-spline trajectory representations \cite{li2023continuous,li2025embedding}, and learning-based models \cite{yang2022robust, kao2023viunet, arjmandi2024deepcovpg, zheng2023enhanced, huang2024novel, he2024multi, he2025graph, wang2025neural}.
EKF-based methods \cite{li2018accurate,cao2020accurate} fuse ranging and inertial measurements via the Kalman filtering framework, while the loosely-coupled integration scheme and Gaussian noise assumption may result in suboptimal localization performance.
Nonlinear optimization–based approaches \cite{jiang2023efficient,hashim2023nonlinear} can effectively integrate heterogeneous sensing modalities. Nonetheless, these methods are sensitive to the quality of initial conditions. Poor initialization or significant range errors induced by non-line-of-sight (NLOS) can cause the optimizer to fail to converge or to become trapped in local minima.
B-spline approaches \cite{li2023continuous,li2025embedding} support interpolation at any time instant while enforcing continuous motion-model constraints, making them well suited for scenarios with high ranging noise or sparse observations. Yet, they also exhibit strong sensitivity to initialization, and necessitate extensive hyperparameter tuning such as knot density and sliding window size. Learned methods \cite{yang2022robust, kao2023viunet, arjmandi2024deepcovpg, zheng2023enhanced, huang2024novel, he2024multi, he2025graph, wang2025neural} are capable of modeling the non-Gaussian distribution of UWB ranging errors and exploiting multimodal information in a tightly-coupled manner, leading to improved localization performance. VIUNet \cite{kao2023viunet} features a CNN-LSTM module for visual-inertial odometry and a MLP module for UWB absolute localization, respectively, achieving visual-inertial-ranging fusion for single-robot single-tag localization. MLP-, CNN-, and LSTM-based \cite{yang2022robust, kao2023viunet} networks struggle to handle dynamic UWB nodes, whereas graph neural network (GNN)-based methods \cite{zheng2023enhanced, huang2024novel, he2024multi, he2025graph, wang2025neural} stand out for UWB localization due to their dynamic topological structure and superior feature aggregation capabilities. NRIO \cite{wang2025neural} models the UWB localization subsystem using a graph neural network and couples it with GRU-driven inertial odometry, enabling robust, high-accuracy multi-tag localization on a single robot.

\subsubsection{Multi-robot Positioning}
In a multi-robot localization system, each robot estimates the positions of its neighboring robots in its own local coordinate frame.
Due to the under-determined nature of the multi-robot relative localization problem, inter-robot distance measurements alone are typically insufficient to maintain observability, necessitating auxiliary bearing information from Angle-of-Arrival (AoA) \cite{zhao2025robust,liang20253} or vision-based methods \cite{xun2023crepes,olson2011apriltag,yan2019active,xu2020decentralized,xu2022omni,xu2024d,wang2025mr}. Visual detections deliver explicit and highly informative bearing observations but lack identity labels by nature, rendering the resulting data-association problem particularly challenging. Consequently, prior work has investigated both hardware-encoded schemes \cite{xun2023crepes,olson2011apriltag,yan2019active} and algorithmic matching-based solutions \cite{xu2020decentralized,xu2022omni,xu2024d,wang2025mr} to robustly resolve the measurement correspondence ambiguity.
In this work, we focus on the most challenging scenario, realizing multi-robot relative localization relying solely on anonymous visual detections and inter-robot ranging.
As a representative approach within this line of research, Mr. Virgil \cite{wang2025mr} adopts GNN and Sinkhorn algorithm for data association, with a learnable match front-end and a differentiable PGO back-end, achieving robust multi-robot tracking.
Notably, in multi-robot configurations equipped with multiple antennas per robot, explicit bearing observations are optional. MURP \cite{fishberg2024murp} develops a multi-tag relative localization system, achieving low communication cost with pure range measurements.

\subsection{Unified Architecture}

Methods built upon the above specialized architectures can achieve excellent performance for their designated localization problems. However, their task-specific problem formulations and network designs limit their applicability and transferability to broader localization settings.

Within the image segmentation literature, extensive studies \cite{cheng2021per,cheng2022masked, jain2023oneformer} have explored unified frameworks that jointly address semantic, instance, and panoptic segmentation, and such universal models have demonstrated superior performance compared with specialized architectures on individual segmentation tasks. Mask2Former \cite{cheng2022masked} requires individual training on specific tasks to achieve optimal performance, while OneFormer \cite{jain2023oneformer} constructs a unified dataset to jointly train a single and universal model.
In the field of simultaneous localization and mapping (SLAM), SLAM-Former \cite{yuan2025slam} unifies the visual SLAM workflow (tracking and incremental mapping front-end, optimization back-end) within one Transformer, achieving online dense monocular SLAM with global consistency.

However, within the domain of position tracking based on anonymous bearing and ranging information, no existing approach provides a unified solution capable of addressing the full spectrum of scenarios, including both single and multiple robots as well as single-tag and multi-tag configurations.

\section{Formulation of Unified Position Tracking}

In this section, we begin by defining the notations and symbols used throughout this paper. 
Subsequently, we discuss the most complex but general localization case, namely the multi-robot multi-tag localization task.
Finally, we demonstrate how diverse position tracking tasks can be embedded into a unified formulation by expressing them as special cases of the general multi-robot multi-tag localization problem.
  
\subsection{Notation}

We denote the world (or map) coordinate frame as $\it W$, the static anchor as $\it A$, the robot base frame as $\it B$, and the tag attached to the robot as $\it T$. The position and orientation of the robot base frame with respect to the world frame are written as $\mathbf {t}_{B}^{W}\in \mathbb{R}^3$ and $\mathbf {R}_{B}^{W}\in \mathbb{R}^{3\times3}$ respectively, where the rotation matrix is parameterized by quaternion in our implementation. 
The 6-DoF pose of frame $\it B$ relative to the frame $\it W$ is denoted by $\mathbf {P}_{B}^{W}$, which is equivalent to the transformation matrix 
$\begin{bmatrix}
\mathbf{R}_{B}^{W} & \mathbf{t}_{B}^{W}\\
 0 & \mathbf{1}
\end{bmatrix}$.
The estimated quantities are represented by $\hat{(\cdot)}$. The symbol $\parallel \cdot \parallel$ denotes the Euclidean norm of a vector.
Let $\mathit{N}_A$, $\mathit{N}_R$, $\mathit{N}_T$, $\mathit{N}_C$, and $\mathit{N}_W$ denote the numbers of UWB anchors, robots, UWB tags per robot, max camera bearing detections, and time window size in multi-frame matching, respectively. 
Considering the possibility of erroneous observations, $\mathit{N}_C$ is set larger than $\mathit{N}_R -1$.
$\mathcal{S_R}=\{\mathit{R_i} | 1\le i \le \mathit{N}_R\}$ is the set of robot swarms.

\begin{table}[tp]
\caption{Unified multi-robot multi-tag (MM) formulation for four typical positioning tasks. \# R, \# T, \# A refer to the number of robots, tags per robot, and anchors, respectively.}
\label{table:uni-formulation}
\renewcommand\arraystretch{1.1}
\centering
\begin{tabular}{cc|ccc|cccc}
\hline
\multicolumn{2}{c|}{\multirow{3}{*}{\textbf{Task}}}                                                                            & \multicolumn{3}{c|}{\textbf{Original Setting}}                        & \multicolumn{4}{c}{\textbf{Unified MM Formulation}}                 \\ \cline{3-9} 
\multicolumn{2}{c|}{}                                                                                                          & \multirow{2}{*}{\# R} & \multirow{2}{*}{\# T} & \multirow{2}{*}{\# A} & \multicolumn{2}{c|}{Reference}      & \multicolumn{2}{c}{Estimated} \\ \cline{6-9} 
\multicolumn{2}{c|}{}                                                                                                          &                       &                       &                       & \# R   & \multicolumn{1}{c|}{\# T}  & \# R          & \# T          \\ \hline
\multirow{2}{*}{\begin{tabular}[c]{@{}c@{}}Single\\ Robot\end{tabular}} & \begin{tabular}[c]{@{}c@{}}Single\\ Tag\end{tabular} & 1                     & 1                     & $\mathit{N}_A$                    & 1      & \multicolumn{1}{c|}{$\mathit{N}_A$}    & 1             & 1             \\ \cline{2-9} 
                                                                        & \begin{tabular}[c]{@{}c@{}}Multi\\ Tag\end{tabular}  & 1                     & $\mathit{N}_T$                    & $\mathit{N}_A$                    & 1      & \multicolumn{1}{c|}{$\mathit{N}_A$}    & 1             & $\mathit{N}_T$            \\ \hline
\multirow{2}{*}{\begin{tabular}[c]{@{}c@{}}Multi\\ Robot\end{tabular}}  & \begin{tabular}[c]{@{}c@{}}Single\\ Tag\end{tabular} & $\mathit{N}_R$                    & 1                     & 0                     & 1      & \multicolumn{1}{c|}{1}     & $\mathit{N}_R$-1          & 1             \\ \cline{2-9} 
                                                                        & \begin{tabular}[c]{@{}c@{}}Multi\\ Tag\end{tabular}  & $\mathit{N}_R$                    & $\mathit{N}_T$                    & 0                     & 1      & \multicolumn{1}{c|}{$\mathit{N}_T$}    & $\mathit{N}_R$-1          & $\mathit{N}_T$            \\ \hline
\end{tabular}
\end{table}

\subsection{Multi-robot Multi-tag Positioning}

In this setting, the objective is to predict the relative poses of all neighbour robots 
$\{\mathbf {\hat P}_{B_i}^{B_r} = 
\begin{bmatrix}
\mathbf {\hat R}_{B_i}^{B_r} & \mathbf {\hat t}_{B_i}^{B_r}\\
 0 & \mathbf{1}
\end{bmatrix} |i,r \in \mathcal{S_R},i \ne r\}$
with respect to a reference robot $\mathit{R_r}$ , which can be any robot in the group. 
There are three primary types of constraints: extrinsic constraints, distance constraints, and bearing constraints, which are described as:
\begin{equation}
\label{con:ori-mm}
\begin{aligned}
\mathbf{t}_{T_{i,m}}^{B_r} &= \mathbf{\hat R}_{B_i}^{B_r} \mathbf{t}_{T_{i,m}}^{B_i} + \mathbf{\hat t}_{B_i}^{B_r} + \, \boldsymbol{\eta}_e\\
\textit{d}^{i,m}_{j,n} &= \parallel \mathbf{t}_{T_{i,m}}^{B_r}-\mathbf{t}_{T_{j,n}}^{B_r} \parallel + \, \eta_d \\
\mathbf{b}_{B_{i}}^{B_r} &= \frac{\mathbf{\hat t}_{B_i}^{B_r}}{\parallel \mathbf{\hat t}_{B_i}^{B_r} \parallel} + \, \boldsymbol{\eta}_b
\end{aligned}
\end{equation}
The relative position of the $m$-th tag attached to $\mathit{R_i}$ is denoted as $\mathbf{t}_{T_{i,m}}^{B_r}$, which is calculated by the tag extrinsics $\mathbf{t}_{T_{i,m}}^{B_i}$ and the predicted pose. The $\textit{d}^{i,m}_{j,n}$ refers to the pairwise distance between the $m$-th tag on the $i$-th robot and the $n$-th tag on the $j$-th robot. The $\mathbf{b}_{B_{i}}^{B_r}$ denotes the bearing vector of robot $R_i$ observed by robot $R_r$ in its local frame. The inter-robot bearing vectors can be produced through either rule-based hard matching or learning-based soft assignment. We evaluate the performance differences between distinct matching strategies in the experimental analysis in Appendix \ref{subsec:app-func-valid}. The $\boldsymbol{\eta}_e$, $\eta_d$, and $\boldsymbol{\eta}_b$ correspond to the bias and noise terms associated with the extrinsic, distance, and bearing measurements, respectively.

\subsection{Unified Formulation}
\label{subsec:uni-form}

The single-tag and non-bearing settings still fall within the general category of multi-tag with bearing formulation we discussed before, differing only by the degeneration of their respective constraints.
Thus, forming a unified formulation requires addressing two key issues: (1) converting single-robot global localization in the map frame into multi-robot relative localization, and (2) incorporating the information from static anchors fixed in the map frame. 

To overcome these issues, we reinterpret the map frame in single-robot localization as a virtual reference robot. The anchors defined in the map frame are considered as tags mounted on the reference robot, whose map-frame locations serve as the extrinsic parameters of the UWB tags.
Through this reformulation, the single-robot positioning task is converted into a multi-robot relative positioning problem involving a reference robot (with multiple tags) and a target robot to be estimated.
Tab. \ref{table:uni-formulation} summarizes the configurations of the four representative localization tasks, including robots, anchors, and tags in each scenario before and after the reformulation.

\begin{figure}[tp]
    \centering
    \includegraphics[width=.48\textwidth]{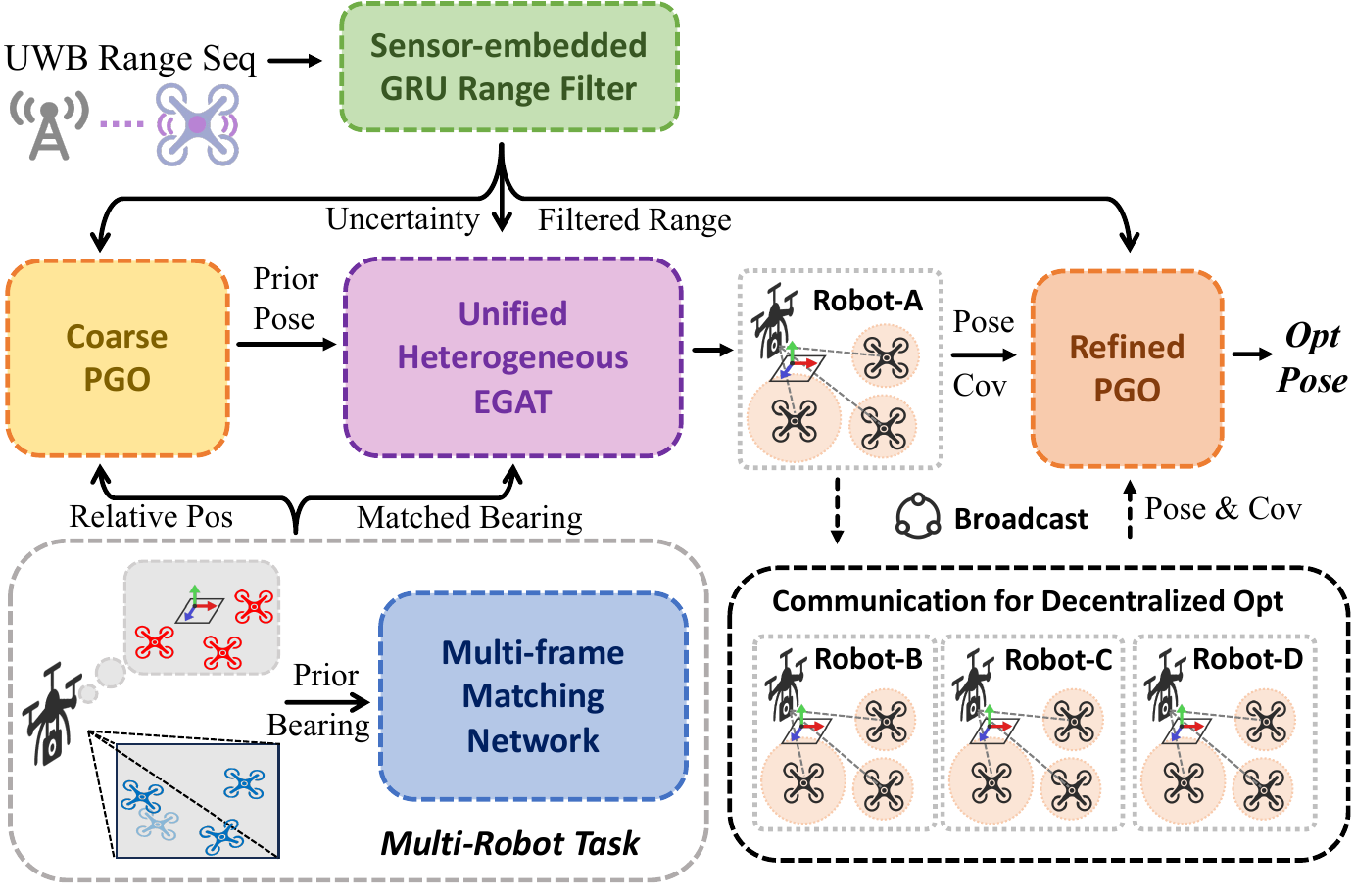}
    \caption{The framework of our proposed \textbf{AnyAmber}, unified anonymous bearing and range based position tracking system.}
    \label{fig:pipeline}
\end{figure}

\section{Network Architecture}

Our system overview is presented in Fig. \ref{fig:pipeline}, which consists of a GRU range filter network, a multi-frame matching module, the key unified heterogeneous EGAT network and two differentiable PGO blocks.

Given a sequence of UWB ranging inputs and the corresponding sensor feature embeddings, the Sensor-Embedded GRU performs filtering and ranging uncertainty estimation, and the resulting features are fed into the downstream Coarse PGO (CPGO), EGAT, and Refined PGO (RPGO) modules.
In multi-robot scenarios with visual detections, the matched bearing information and relative positions are extracted from anonymous visual observations by a GNN-based multi-frame matching network.
With the comprehensive inputs of filtered ranges, matched bearings and pose priors, the core EGAT estimates both the poses and covariances of the robots.
The CPGO and RPGO modules perform prior pose correction and output pose refinement at different stages, respectively.

Our system operates in a decentralized manner when applied to multi-robot settings, where robots communicate only by broadcasting predicted poses and associated covariances after EGAT, and subsequently perform RPGO independently, leading to minimal communication overhead.
In the following sections, we provide a detailed introduction to the design and functionality of each submodule.

\begin{figure}[tp]
\setlength{\belowcaptionskip}{-0.6cm}
    \centering
    \includegraphics[width=.4\textwidth]{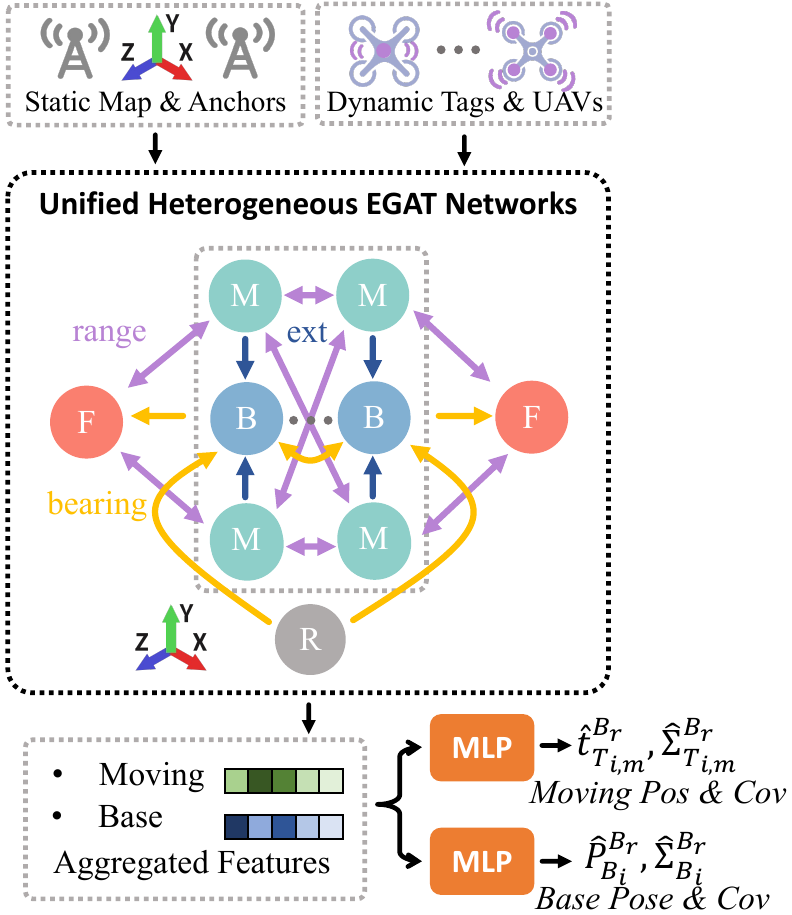}
    \caption{The heterogeneous unified EGAT localization network, incorporating four node types (Ref, Base, Fixed, Moving) and three distinct edge types (ext, range, bearing). The GNN-aggregated moving and base features are utilized to estimate the corresponding pose and covariance.}
    \label{fig:egat}
\end{figure}

\subsection{Heterogeneous Unified EGAT Localization Network}
\label{subsec:egat}
Building upon the unified multi-robot multi-tag formulation introduced in Section \ref{subsec:uni-form}, we propose a heterogeneous graph neural network with four node types and three edge types, with details provided in Appendix \ref{subsec:app-egat}.
Different from the approach in \cite{wang2025neural}, which employs a standard GAT network with all information stored in node features, we follow the approach in \cite{monninger2023scene} and adopt an EGAT network. In this design, the estimated quantities are represented as nodes, while geometric constraints such as range and bearing are encoded as edges, providing a more structured and interpretable problem formulation.
The network architecture is illustrated in Fig. \ref{fig:egat}.

\subsubsection{Graph Nodes} 
As outlined before, we unify the map frame in the single-robot scenario with the reference robot in the multi-robot scenario, which is denoted here as \textit{Ref}. Here the \textit{Ref} feature simply represents the identity pose $\mathbf P_{B_r}^{B_r}$ expressed in its own reference frame. 
Similarly, the static anchors are unified with the tags on the reference robot, denoted as \textit{Fixed}, with the feature $\mathbf t_{T_{r,n}}^{B_r}$ representing the tag positions.
The \textit{Base} node denotes the pose $\mathbf {\hat P}_{B_i}^{B_r}$ of the robot being estimated, while the \textit{Moving} node corresponds to the position $\mathbf {\hat t}_{T_{i,m}}^{B_r}$ of the tag attached to it. 
It is noteworthy that \textit{Ref} and \textit{Base} nodes, as well as \textit{Fixed} and \textit{Moving} nodes, could theoretically be further merged into a more unified architecture. We retain their distinction because reference nodes and estimated nodes may vary in confidence and feature aggregation behaviors. Representing them as separate node types yields superior performance.

\subsubsection{Graph Edges} 
The unified localization formulation incorporates three constraint types: extrinsic, bearing, and range constraints, which motivate the three corresponding edge types in the GNN. 
In bearing observations, we typically establish a bearing edge from the reference robot to the target robot, namely \textit{Ref}-to-\textit{Base}. When a bearing observation exists between neighboring robots, a \textit{Base}-to-\textit{Base} edge can be derived. Similarly, a \textit{Base}-to-\textit{Fixed} edge is derived when a robot observes static anchors.
For range measurements, \textit{range} edges are subdivided into \textit{Moving}-to-\textit{Fixed}, \textit{Fixed}-to-\textit{Moving}, and \textit{Moving}-to-\textit{Moving} types based on whether the range node corresponds to an estimated node, and the weights of these range edges are mutually independent.

\subsubsection{Feature Aggregation}
Following the same message aggregation mechanism in \cite{monninger2023scene}, we finally extract the node features of \textit{Moving} and \textit{Base} to predict their poses and associated uncertainties, with details presented in Appendix \ref{subsec:app-egat}.

\begin{figure}[tp]
    \setlength{\belowcaptionskip}{-0.4cm}
    \centering
    \includegraphics[scale=0.47]{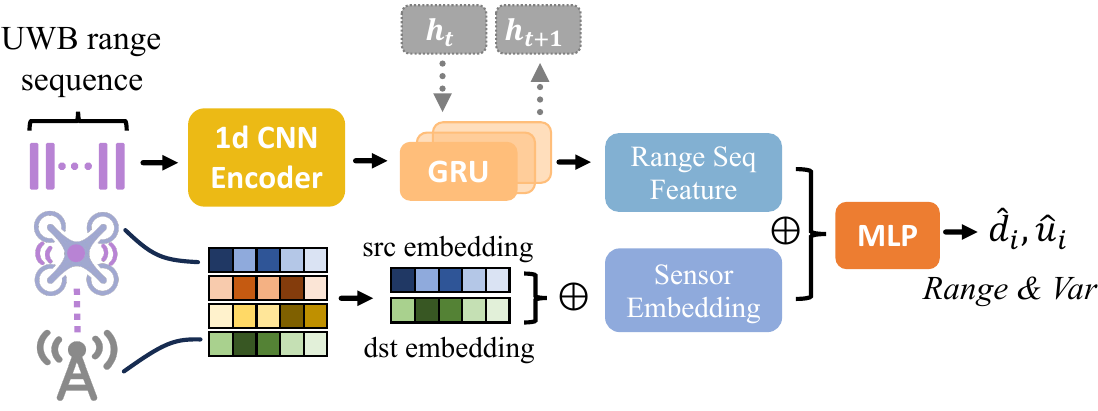}
    \caption{The Sensor-Embedded GRU range filter network.}
    \label{fig:gru}
\end{figure}

\subsection{Sensor-Embedded GRU Network} 
It is widely recognized that UWB ranging can be impaired by noise and systematic bias, often caused by antenna deployment and inter-node signal interference, giving rise to a non-zero-mean, non-Gaussian ranging pattern \cite{jiang2023efficient, nguyen2018robust}. More critically, the noise and bias are typically time-varying and node-pair-dependent, which implies that a generic filter with fixed parameters often yields suboptimal denoising performance.

To tackle this issue, a sensor-embedding–augmented GRU network is introduced to facilitate tailored and adaptive calibration, consisting of a standard GRU module to capture the general temporal ranging pattern for effective suppression of ranging noise during pretraining, and a sensor embedding table to model the scenario-specific pair-wise sensor characteristics for sensor-coupled ranging biases correction during finetuning.
A batch of UWB ranging measurements within a given window is first processed by a 1D CNN to extract temporal ranging features. Subsequently, a three-layer GRU network integrates the historical hidden representations with the current ranging features to generate more expressive range embeddings. To model the distinct characteristics of each sensor pair, the unique IDs of the transmitting and receiving sensors are used to fetch their node embeddings from a pre-trained embedding table. The embeddings from sensor pairs and range sequences are then added together to generate the pairwise sensor embeddings, followed by the MLP to predict the filtered range and its corresponding variance.

\subsection{Multi-frame Matching Network} 
\begin{figure}[tp]
    \setlength{\belowcaptionskip}{-0.4cm}
    \centering
    \includegraphics[width=.48\textwidth]{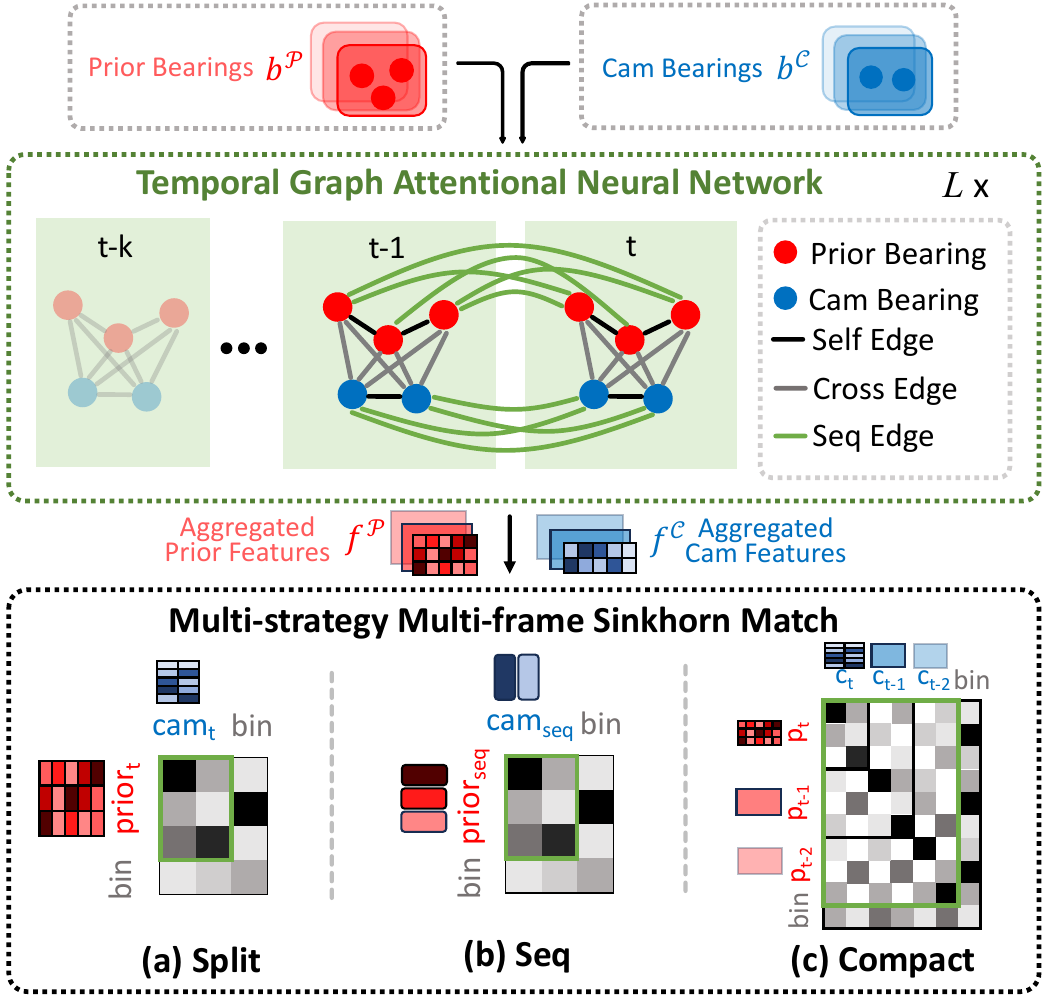}
    \caption{Multi-frame temporal matching network.}
    \label{fig:matchnet-zip}
\end{figure}

Motivated by the research about learnable multi-robot matching \cite{wang2025neural}, we solve the data association problem between ID-aware priors and anonymous visual detections using graph-based approach. To effectively capture temporal dependencies, we extend the single-frame matching paradigm to a multi-frame setting by introducing temporal GNN-based networks (encoder module) and multi-frame Sinkhorn partial assignments (decoder and matching module).
The architecture of our graph match net is illustrated in Fig. \ref{fig:matchnet-zip}.

\subsubsection{Temporal Attentional Graph Neural Network} 

The initial node features of priors $\mathcal{P}$ and camera detections $\mathcal{C}$ are represented by their high-dimensional embeddings $^{(0)}\textbf{f} _{i}$ projected from raw bearings via a shared MLP.
We design three distinct types of cross-attention edges in the GNN: self-edges, cross-edges, and sequential-edges, enabling a more profound integration of the multi-robot formation characteristics (self-edge), the similarity cues from candidate bearing measurements (cross-edge), and the temporal context (seq-edge).
The lifted node embeddings undergo three successive aggregation stages: (1) self-edge aggregation with nodes from the same set at the same time step, (2) cross-edge aggregation with nodes from the other set at the same time step, and (3) sequential aggregation with nodes from different time steps within the same set.
Note that residual connections are utilized at both intra-layer and inter-layer levels to facilitate feature propagation.

\begin{equation}
\begin{aligned}
^{(l)}\textbf{f}^{self}_{i} = ^{(l)}\textbf{f} _{i} + ^{(l)}\mathit{f}_{self}([^{(l)}\textbf{f} _{i} \, || \, ^{(l)}\textbf{m}_{\varepsilon_{self}}]) \\
^{l}\textbf{f}^{cross}_{i} = ^{(l)}\textbf{f}^{self}_{i} + ^{(l)}\mathit{f}_{cross}([^{(l)}\textbf{f}^{self}_{i} \, || \, ^{(l)}\textbf{m}_{\varepsilon_{cross}}]) \\
^{(l+1)}\textbf{f} _{i} = ^{(l)}\textbf{f}^{cross}_{i} + ^{(l)}\mathit{f}_{seq}([^{(l)}\textbf{f}^{cross}_{i} \, || \, ^{(l)}\textbf{m}_{\varepsilon_{seq}}])
\end{aligned}
\end{equation}
where $||$ denotes concatenation. $^{(l)}\textbf{f} _{i}$ is the bearing embedding of layer $l$. The message aggregation along self-edge $\varepsilon_{self}$, cross-edge $\varepsilon_{cross}$ and seq-edge $\varepsilon_{seq}$ are represented by $^{(l)}\textbf{m}_{\varepsilon_{self}}$, $^{(l)}\textbf{m}_{\varepsilon_{cross}}$ and $^{(l)}\textbf{m}_{\varepsilon_{seq}}$ respectively. $^{(l)}\mathit{f}_{self}$, $^{(l)}\mathit{f}_{cross}$ and $^{(l)}\mathit{f}_{seq}$ are MLPs to merge attentional features and own embeddings, ensuring that the feature dimensions remain consistent before and after the aggregation.

After extensive message passing through multiple layers of the GNN, we obtain two bearing feature sets $\textbf{f}^{\mathcal{P}}$ and $\textbf{f}^{\mathcal{C}}$, with enhanced spatiotemporal representation capabilities.

\subsubsection{Multi-frame Partial Assignment} 
We introduce three distinct strategies for performing multi-frame partial assignment, namely $\textit{Split}$, $\textit{Sequential}$ and $\textit{Compact}$.
The $\textit{Split}$ method divides the multi-frame bearing set into individual frames and performs frame-to-frame matching between the prior set and the camera set.
The $\textit{Sequential}$ method formulates multi-frame matching as a two-stage sequential matching problem. 
The $\textit{Compact}$ method constructs a single score matrix $^{k}\mathbf S\in \mathbb{R}^{(N_R*N_W)\times (N_C*N_W)}$ by stacking bearing features from all frames, enabling joint iterative optimization over the entire sequence in a single step.
More details of these three assignment algorithms can be found in Appendix \ref{subsec:app-sinkhorn}.



\subsection{Hierarchical Differentiable PGO}
To enhance the overall position tracking precision, we adopt a hierarchical differentiable PGO architecture composed of CPGO and RPGO. These two components share identical constraint types, but operate on different input data. There are four major constraints involved in PGO: pose prior constraint, extrinsic constraint, UWB ranging constraint and mutual state constraint.
The CPGO module integrates the previous frame’s predictions with the current frame observations to provide a refined pose prior for the EGAT localization network. Subsequently, the RPGO module takes the pose predictions and uncertainty estimates generated by the EGAT network to optimize the final pose output.

In our implementation, we leverage Theseus \cite{pineda2022theseus} for differentiable nonlinear optimization. The gradients backpropagated from this optimization backend facilitate the EGAT network's learning of predictive uncertainties.

\section{Training Settings}

\subsection{Loss Functions}
To jointly improve the performance of matching, filtering, and localization, we define a composite loss function to supervise the network training:
\begin{equation}
\begin{aligned}
\mathcal{L} = \mathcal{L}_{Match} + \lambda_p \mathcal{L}_{Pose} + \lambda_r \mathcal{L}_{Range}
\end{aligned}
\end{equation}
where $\lambda_r$ and $\lambda_p$ are weighting factors for different loss terms.

\subsubsection{Match Loss} 
Following $N_I$ Sinkhorn iterations, we obtain the score matrix $_{(N_I)}\bar{\mathbf S}$, which may represent frame-to-frame, sequence-to-sequence or compact matching. Suppose the matrix has $N+1$ rows and $M+1$ columns.
We supervise the specific item in the iterated score matrix:
\begin{equation}
\begin{aligned}
\mathcal{L}_{Match} = -\sum_{(i,j)\in \pi} {_{(N_I)}\bar{\mathbf S}_{i,j}} - \sum_{i\in \mu} {_{(N_I)}\bar{\mathbf S}_{i,M}} - \sum_{j\in \upsilon} {_{(N_I)}\bar{\mathbf S}_{N,j}}
\end{aligned}
\end{equation}
where $\pi$ is the set of matching bearings, $\mu$ and $\upsilon$ denote the set of unmatched candidates, derived from the ground truth labels. The first term encourages higher scores for correct correspondences, while the latter two terms promote assigning incorrect matches to the dustbin row and column.

\subsubsection{Pose Loss} 
Pose supervision is achieved through two losses: a maximum-likelihood (ML) loss on the EGAT-predicted poses and variances, and a Mean Squared Error (MSE) loss on the RPGO-refined poses, whose gradients are further propagated backward to facilitate end-to-end training of the upstream network.
\begin{equation}
\begin{split}
\mathcal{L}_{Pose} = & ^{ML}\mathcal{L}_{Pose} + \lambda_{m} \cdot ^{MSE}\mathcal{L}_{Pose} \\
^{ML}\mathcal{L}_{Pose} = &\frac{1}{N_S} \sum_{i\in \mathcal{S}} (\lambda_{d}log(det(\mathbf{\hat{\Sigma}}_i))\\
                    &+ (\mathbf{e}_i - \mathbf{\hat{e}}_i)^T \, \mathbf{\hat{\Sigma}}^{{-1}}_{i} \, (\mathbf{e}_i - \mathbf{\hat{e}}_i)) \\
^{MSE}\mathcal{L}_{Pose} = &\frac{1}{N_S} \sum_{i\in \mathcal{S}} (\parallel \mathbf{t}_i - \mathbf{\hat{t}}_i \parallel^2 
                     + \lambda_{q} \parallel \mathbf{q}_i - \mathbf{\hat{q}}_i \parallel^2 )  
\end{split}
\end{equation}
where $\mathcal{S}$ is the set of robots to be estimated and $N_S$ is the number of variables in this set. $\mathbf{\hat{e}}_i \in \mathbb{R}^3$ represents the predicted state of the $i$-th robot, specifically its estimated position or Euler angles (converted from quaternions). While $\mathbf{e}_i$ and $\mathbf{\hat{\Sigma}}_i \in \mathbb{R}^{3\times 3}$ are its corresponding ground truth and predicted covariance matrix.
For the MSE loss term, the $\mathbf{\hat{t}}_i$ and $\mathbf{\hat{q}}_i$ refer to the optimized position and quaternion from the RPGO module. $\lambda_{d}$, $\lambda_{m}$ and $\lambda_{q}$ are the coefficients that balance the loss weights of the determinant term, the MSE term and the quaternion term.

\subsubsection{Range Loss} 
To enable the GRU network to achieve improved range filtering performance and uncertainty estimation, we adopt the ML loss, but reduce it to a one-dimensional form.
\begin{equation}
\begin{split}
\mathcal{L}_{Range} = \frac{1}{N_D} \sum_{i\in \mathcal{D}} (\lambda_{d}log(\hat u_i) + \frac{( d_i - {\hat{d}}_i )^2}{\hat u_i})\\
\end{split}
\end{equation}
where $\mathcal{D}$ is the set of distance measurements to be filtered and $N_D$ is the number of ranging in this set.
$\hat d_i$, $\hat u_i$ and $d_i$ are the filtered distance, uncertainty prediction and ground truth distance, respectively.

\begin{table*}[tp]
\caption{Multi-scene multi-task positioning RMSE (m). All test scenarios are strictly unseen during pre-training. Test scenarios marked with \textit{*} indicate that both the scenario itself and its scenario type have never been encountered during pre-training. The simulation scenarios are denoted by the \textit{sim-} prefix. The best positioning result is highlighted in bold, the second-best is marked with underlining, the - indicates that the method is not applicable to the corresponding localization task.}
\label{table:position-accuracy}
\renewcommand\arraystretch{1.2}
\setlength{\tabcolsep}{3.5pt}
\centering
\begin{tabular}{cccc|cccccccc}
\hline
\multicolumn{2}{c}{\multirow{3}{*}{\textbf{Method}}}                                                                                               & \multirow{3}{*}{\textbf{Learned}} & \multirow{3}{*}{\textbf{Unified}} & \multicolumn{4}{c}{\textbf{Single-Robot}}                                        & \multicolumn{4}{c}{\textbf{Multi-Robot}}                                                                                                                                                                                                                                \\ \cline{5-12} 
\multicolumn{2}{c}{}                                                                                                                               &                                   &                                   & \multicolumn{2}{c}{\textbf{Single-Tag}} & \multicolumn{2}{c}{\textbf{Multi-Tag}} & \multicolumn{2}{c}{\textbf{Single-Tag}}                                                                                            & \multicolumn{2}{c}{\textbf{Multi-Tag}}                                                                                             \\ \cline{5-12} 
\multicolumn{2}{c}{}                                                                                                                               &                                   &                                   & sim-circle         & square*\cite{kao2023viunet}            & \begin{tabular}[c]{@{}c@{}}tunnel\\ drive\end{tabular}       & \begin{tabular}[c]{@{}c@{}}tunnel\\ fly\end{tabular}        & \begin{tabular}[c]{@{}c@{}}lab-los*\cite{wang2025mr}\\ (w. bearing)\end{tabular} & \begin{tabular}[c]{@{}c@{}}lab-nlos*\cite{wang2025mr}\\ (w. bearing)\end{tabular} & \begin{tabular}[c]{@{}c@{}}sim-spiral\\ (w. bearing)\end{tabular} & \begin{tabular}[c]{@{}c@{}}plane*\cite{fishberg2024murp}\\ (wo. bearing)\end{tabular} \\ \hline
\multirow{3}{*}{\textbf{\begin{tabular}[c]{@{}c@{}}Individual\\ Training\end{tabular}}} & \begin{tabular}[c]{@{}c@{}}VIUNet\cite{kao2023viunet}\\ (SS)\end{tabular}    & \checkmark                               & \ding{55}                                & 1.08               & {\underline{0.12}}         & -                  & -                 & -                                                               & -                                                                & -                                                                 & -                                                              \\ \cline{2-12} 
                                                                                        & \begin{tabular}[c]{@{}c@{}}NRIO\cite{wang2025neural}\\ (SM)\end{tabular}      & \checkmark                               & \ding{55}                                & -                  & -                  & {\underline{0.18}}         & {\underline{0.26}}        & -                                                               & -                                                                & -                                                                 & -                                                              \\ \cline{2-12} 
                                                                                        & \begin{tabular}[c]{@{}c@{}}Mr.Virgil\cite{wang2025mr}\\ (MS)\end{tabular} & \checkmark                               & \ding{55}                                & -                  & -                  & -                  & -                 & {\underline{0.63}}                                                      & \textbf{0.14}                                                    & -                                                                 & -                                                              \\ \hline
\multirow{2}{*}{\textbf{\begin{tabular}[c]{@{}c@{}}Optimization\\ -based\end{tabular}}}  & \begin{tabular}[c]{@{}c@{}}MURP\cite{fishberg2024murp}\\ (MM)\end{tabular}      & \ding{55}                                & \ding{55}                                & -                  & -                  & -                  & -                 & -                                                               & -                                                                & -                                                                 & {\underline{1.44}}                                                     \\ \cline{2-12} 
                                                                                        & PGO                                                      & \ding{55}                                & \checkmark                               & {\underline{0.63}}         & 0.14               & 0.59               & 0.44              & 0.77                                                            & 0.17                                                             & 5.66                                                             & 1.78                                                           \\ \hline
\textbf{\begin{tabular}[c]{@{}c@{}}Joint\\ Training\end{tabular}}                       & \textbf{AnyAmber}                                        & \checkmark                               & \checkmark                               & \textbf{0.17}      & \textbf{0.04}      & \textbf{0.15}      & \textbf{0.11}     & \textbf{0.09}                                                   & {\underline{0.15}}                                                       & \textbf{0.11}                                                     & \textbf{0.36}                                                  \\ \hline
\end{tabular}
\end{table*}

\subsection{Joint Pretraining and Scene-specific Finetuning}

During pretraining, for efficiency considerations, the CPGO and RPGO modules are omitted. The EGAT localization network, the sensor-agnostic GRU network, and the GNN matching network are pretrained independently, forming transferable general knowledge for localization, filtering, and matching.
Although various localization tasks are unified under a multi-robot multi-tag formulation, some degrees of freedom are still underdetermined in specific settings, such as the ambiguity of orientation in single-tag scenarios without bearing information. To facilitate joint pretraining, we adopt a unified output representation and apply a state mask to supervise the estimated states selectively.

Different localization tasks often involve diverse scene-dependent factors, such as varying UWB node distributions, heterogeneous sensor configurations, and differing noise characteristics. As a result, it is necessary to perform fine-tuning tailored to the characteristics of each individual scenario.

In the fine-tuning stage, we augment the GRU network with sensor embeddings to learn scene-specific ranging noise properties. Furthermore, the CPGO and RPGO modules are integrated into the end-to-end training framework, allowing gradients from the optimization procedure to backpropagate and supervise the EGAT network in covariance learning.

The model has encapsulated generalized localization knowledge after pretraining, enabling effective fine-tuning for specific scenes with limited data. In practice, we observe that fine-tuning on merely one trajectory yields substantial performance improvements in the target scenario.

\section{Experiment}

\subsection{Experimental Settings and Datasets} 
We developed a comprehensive multi-scenario localization dataset comprising both simulation and real-world experiments. The dataset spans diverse conditions, including varying numbers of robots, anchors, and UWB tags, alongside diverse robot trajectories and environmental influences. It contains 113 distinct scenes, 317 trajectories, and 210589 sampled frames, among which 108 scenes and 225 trajectories are used for joint pretraining, encompassing configurations with 1 to 16 robots, 0 to 36 anchors, and 1 to 6 tags. The details of our unified dataset and settings are presented in Appendix \ref{subsec:app-data}. 

We evaluate the RMSE localization errors of our method against general and task-specific approaches. Detailed descriptions of the baselines are provided in the Appendix \ref{subsec:app-baseline}.
Since the learning-based baselines are designed with specialized network architectures, we train each of them individually using as much training data as possible on its corresponding task. Our model, by contrast, is first pre-trained on the unified dataset and then fine-tuned in each specific test scenario using only a single trajectory of approximately 150 frames.

\subsection{Multi-scene Position Tracking Accuracy}

The comprehensive experimental results are presented in Tab. \ref{table:position-accuracy}. Representative experimental trajectories are visualized in Fig. \ref{fig:traj-part1}, with additional results available in Appendix \ref{subsec:addi-exp}. Our method not only adapts to various positioning tasks, including single-robot, multi-robot, single-tag, and multi-tag configurations, but also achieves optimal or near-optimal performance in each specific scenario.



\begin{figure*}[htbp]
    \centering
    \includegraphics[scale=0.4]{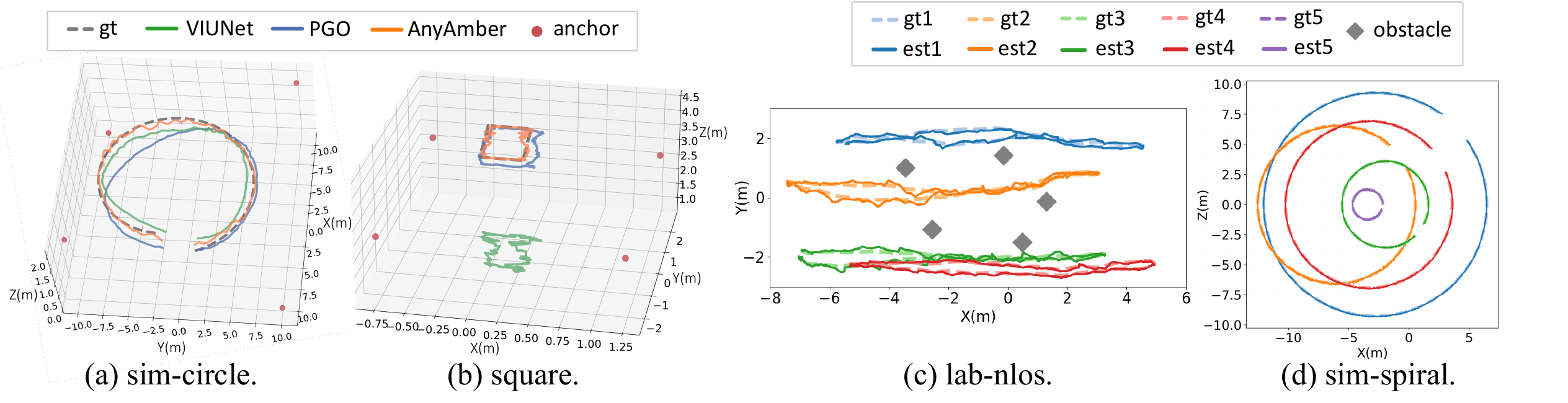}
    \caption{Visualization of representative trajectory estimates in single-robot (a–b) and multi-robot (c–d) scenarios.}
    \label{fig:traj-part1}
\end{figure*}

In the single-robot scenario, the error metrics are calculated in the nominal frame. 
Fig. \ref{fig:traj-part1} (a)(b) compares the trajectories of ground-truth, our method's prediction, and baseline predictions. 
Our model outperforms all learning-based and optimization-based approaches owing to the strong foundation established by the unified and learnable architecture.

For the multi-robot case in Fig. \ref{fig:traj-part1} (c)(d), we visualize only our predictions against the ground truth for clarity, where neighbor robot trajectories are plotted in the world frame by transforming their estimated relative poses using the world pose of the reference robot. The error metrics in Tab. \ref{table:position-accuracy} are calculated in the local frame of the reference robot.
Under the \textit{lab-los} scenario where the robots move at relatively high speeds, Mr. Virgil’s accuracy deteriorates significantly once pseudo odometry is removed, due to its heavy reliance on prior estimates. In contrast, our method limits the localization error to under 10 cm. In the multi-robot multi-tag scenario \textit{plane}, the z-axis is weakly constrained and no bearing measurements are available, making the task considerably challenging. Despite this, our method maintains a relatively low localization error, with errors reduced to 25\% or less of those of the baselines, demonstrating strong robustness.

\begin{table*}[tp]
\caption{Ablation study on system architecture evaluated by localization RMSE (m)}
\label{table:ablation-net}
\renewcommand\arraystretch{1.2}
\setlength{\tabcolsep}{4.5pt}
\centering
\begin{tabular}{cccc|ccccccccc}
\hline
\multirow{3}{*}{\textbf{EGAT}} & \multirow{3}{*}{\textbf{GRU}} & \multirow{3}{*}{\textbf{\begin{tabular}[c]{@{}c@{}}Coarse\\ PGO\end{tabular}}} & \multirow{3}{*}{\textbf{\begin{tabular}[c]{@{}c@{}}Refined\\ PGO\end{tabular}}} & \multicolumn{4}{c}{\textbf{Single-Robot}}                                                                                                               & \multicolumn{4}{c}{\textbf{Multi-Robot}}                                                                                                                                                                                                                                & \multirow{3}{*}{\textbf{Avg.}} \\ \cline{5-12}
                               &                               &                                                                                &                                                                                 & \multicolumn{2}{c}{\textbf{Single-Tag}} & \multicolumn{2}{c}{\textbf{Multi-Tag}}                                                                        & \multicolumn{2}{c}{\textbf{Single-Tag}}                                                                                            & \multicolumn{2}{c}{\textbf{Multi-Tag}}                                                                                             &                                \\ \cline{5-12}
                               &                               &                                                                                &                                                                                 & sim-circle         & square*            & \begin{tabular}[c]{@{}c@{}}tunnel\\ drive\end{tabular} & \begin{tabular}[c]{@{}c@{}}tunnel\\ fly\end{tabular} & \begin{tabular}[c]{@{}c@{}}lab-los*\\ (w. bearing)\end{tabular} & \begin{tabular}[c]{@{}c@{}}lab-nlos*\\ (w. bearing)\end{tabular} & \begin{tabular}[c]{@{}c@{}}sim-spiral\\ (w. bearing)\end{tabular} & \begin{tabular}[c]{@{}c@{}}plane*\\ (wo. bearing)\end{tabular} &                                \\ \hline
\checkmark                            &                               &                                                                                &                                                                                 & 0.44               & \textbf{0.04}      & 0.45                                                   & 0.24                                                 & 0.18                                                            & 0.20                                                             & 0.27                                                              & 3.13                                                           & 0.62                           \\ \hline
\checkmark                            &                               & \checkmark                                                                            & \checkmark                                                                             & 0.28               & \textbf{0.04}      & \textbf{0.15}                                          & \textbf{0.11}                                        & 0.12                                                            & 0.17                                                             & \underline{0.13}                                                              & 1.23                                                           & 0.28                           \\ \hline
\checkmark                            & \checkmark                           &                                                                                & \checkmark                                                                             & \textbf{0.16}      & \textbf{0.04}      & 0.16                                                   & \textbf{0.11}                                        & \textbf{0.09}                                                   & {\underline{0.16}}                                                       & \underline{0.13}                                                     & 0.82                                                           & {\underline{0.21}}                     \\ \hline
\checkmark                            & \checkmark                           & \checkmark                                                                            &                                                                                 & 0.35               & \textbf{0.04}      & 0.21                                                   & 0.13                                                 & 0.37                                                            & 0.27                                                             & 0.20                                                              & {\underline{0.45}}                                                     & 0.25                           \\ \hline
\checkmark                            & \checkmark                           & \checkmark                                                                            & \checkmark                                                                             & {\underline{0.17}}         & \textbf{0.04}      & \textbf{0.15}                                          & \textbf{0.11}                                        & \textbf{0.09}                                                   & \textbf{0.15}                                                    & \textbf{0.11}                                                     & \textbf{0.36}                                                  & \textbf{0.15}                  \\ \hline
\end{tabular}
\end{table*}

\subsection{Ablation Study}
\label{subsec:ablation}
We conduct ablation studies on different network architectural configurations to demonstrate the functionality of each submodule, with quantitative results reported in Tab. \ref{table:ablation-net}.
The Sensor-Embedded GRU can model non-Gaussian UWB ranging characteristics and correct sensor biases effectively, particularly in scenarios with large ranging noise and bias, reducing the mean localization error by over 40\%.
In most scenarios, systems without Coarse PGO can still achieve comparable localization accuracy.
However, in more challenging scenarios, such as \textit{plane}, the correction of prior poses via Coarse PGO provides more accurate initialization, thereby promoting network convergence.
Refined PGO functions as the backend of the system, combining sensor observations with EGAT-derived pose and uncertainty estimates, which substantially enhances localization performance for most localization tasks.

\subsection{Network Generalization Ability Analysis}
In this section, we investigate the generalization capability of the model through experiments under different numbers of training scenes and varying robot counts.

\subsubsection{Training Trajectory Numbers} 
We evaluate the localization performance of pretrained and non-pretrained models in a single-robot, single-tag simulation scenario, fine-tuning each model with different numbers of trajectories.
As illustrated in Fig. \ref{fig:train-traj-num}, the pretrained model, after being fine-tuned on merely one trajectory, generalizes effectively to unseen scenarios and yields significantly lower localization error than the non-pretrained model even when the latter is fine-tuned on five trajectories. This highlights that the joint pretraining process endows the model with transferable multi-scene localization knowledge, substantially alleviating data demands during downstream deployment.

\subsubsection{Robot Numbers} 
\begin{figure}[tp]
    \setlength{\belowcaptionskip}{-0.2cm}
    \centering
    \includegraphics[width=.48\textwidth]{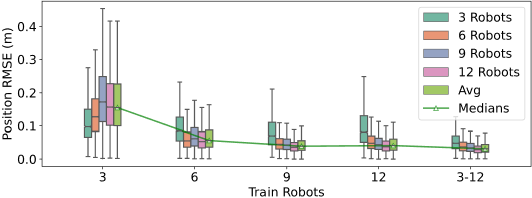}
    \caption{RMSE box plot of varying robot numbers experiment.}
    \label{fig:robot-num-box}
\end{figure}

To evaluate the model’s ability to handle different robot scales, we train models from scratch (without pretraining) on datasets containing 3, 6, 9, 12, and 3–12 robots, and test each model across scenes with varying numbers of robots.
According to the error box plots in Fig. \ref{fig:robot-num-box}, the model trained on the mixed 3–12 robot dataset exhibits strong generalization across test scenarios with varying robot numbers,
and can even surpass the models trained specifically for a fixed robot count on its corresponding scenario.

\subsection{Computational Efficiency and Scalability}

To ensure real‑time inference, we implement the computationally heavy CPGO and RPGO modules in C++ using Ceres Solver, and leverage pybind11 for high‑performance calls from Python during inference. The neural network inference is accelerated via GPU, batch processing, and parallelization. As computation time is summarized in Tab. \ref{table:time}, the system achieves an output rate of at least 20 Hz, even when scaling to 16 robots. We also vary the observation neighborhood in a 16‑robot setup; the matching network runtime remains nearly unchanged (20.94 ms for size 6 vs. 20.97 ms for size 15), highlighting the model’s scalability.

\begin{table}[tp]
\caption{Computational time (ms) of each submodule and the full pipeline under varying robot scales.}
\label{table:time}
\centering
\begin{tabular}{c|ccccc|c}
\hline
\#Robot & MatchNet & GRU & CPGO & EGAT & RPGO & \textbf{Total} \\ \hline
4       & 21.2     & 0.9 & 0.2  & 11.8 & 0.2  & 34.3  \\
8       & 20.7     & 0.9 & 0.6  & 11.5 & 0.5  & 34.3  \\
12      & 21.2     & 0.9 & 1.7  & 11.5 & 1.6  & 36.9  \\
16      & 22.0     & 0.9 & 5.0  & 11.6 & 3.7  & 43.2  \\ \hline
\end{tabular}
\end{table}

\section{Conclusion} 
\label{sec:conclusion}

In this work, we propose a unified modeling formulation and network architecture capable of handling a wide range of anonymous bearing and ranging based localization tasks, including single-robot, multi-robot, single-tag, and multi-tag settings. We construct a comprehensive multi-task dataset for pretraining, integrating simulation and real-world data, self-collected and public datasets. The pretrained model can be adapted to unseen scenarios with only a single fine-tuning trajectory and achieves superior or comparable localization accuracy to task-specific architectures.

Nevertheless, due to the output-level fusion of the matching and localization networks in current methods, positional priors remain indispensable, preventing our approach from scaling toward global localization in a fully prior-free manner. Notably, our architecture readily supports extension to a truly integrated end‑to‑end design (e.g., feeding GRU/MatchNet latent features directly into EGAT). As part of our future efforts, 
we will explore prior‑free global localization by leveraging odometry, multi‑frame observations, and feature‑level end‑to‑end fusion.

\section*{Acknowledgments}
This work was supported by the Fundamental and Interdisciplinary Disciplines Breakthrough Plan of the Ministry of Education of China under Grant No. JYB2025XDXM208, the National Nature Science Foundation of China under Grant No. 62522317 and 62503144, the Joint Funds of the National Natural Science Foundation of China under Grant No. U24A20128, Zhejiang Provincial Natural Science Foundation of China under Grant No. LD25F030001, the Key R\&D Program of Hangzhou under Grant No. 2024SZD1A30.


\bibliographystyle{unsrt}
\bibliography{references}

\clearpage

\appendix

\subsection{Features of Heterogeneous EGAT Localization Network}
\label{subsec:app-egat}

\begin{table}[htbp]
\caption{Three edge types in heterogeneous EGAT.}
\label{table:edge_feat}
\setlength{\tabcolsep}{4.0pt}
\renewcommand\arraystretch{1.6}
\centering
\begin{tabular}{c|cc|c|c}
\hline
\multirow{2}{*}{\textbf{\begin{tabular}[c]{@{}c@{}}Edge\\ Type\end{tabular}}} & \multicolumn{2}{c|}{\textbf{Relation}} & \multirow{2}{*}{\textbf{Feat}} & \multirow{2}{*}{\textbf{Description}}                                                                 \\ \cline{2-3}
                                                                              & \textbf{src}       & \textbf{dst}      &                                &                                                                                                       \\ \hline
ext                                                                           & Moving             & Base              & $\mathbf t_{T_{i,m}}^{B_i}$                              & Tag $m$'s extrinsics of robot $i$                                                                                       \\ \hline
\multirow{3}{*}{bearing}                                                      & Ref                & Base              & $\mathbf b_{B_i}^{B_r}$                              & \begin{tabular}[c]{@{}c@{}}Normalized bearings from\\ ref robot $r$ to predicted robot $i$\end{tabular}       \\ \cline{2-5} 
                                                                              & Base               & Base              & $\mathbf b_{B_j}^{B_i}$                              & \begin{tabular}[c]{@{}c@{}}Normalized bearings from\\ predicted robot $i$ to robot $j$\end{tabular} \\ \cline{2-5} 
                                                                              & Base               & Fixed             & $\mathbf b_{T_{r,n}}^{B_i}$                              & \begin{tabular}[c]{@{}c@{}}Normalized bearings from\\ predicted robot $i$ to anchor $n$\end{tabular}    \\ \hline
\multirow{3}{*}{range}                                                        & Moving             & Fixed             & \multirow{3}{*}{$\textit{d}^{i,m}_{j,n}$}             & \multirow  {3}{*}{\begin{tabular}[c]{@{}c@{}}Range between the tag $m$ on \\ robot $i$ and the tag $n$ on robot $j$\end{tabular}}                                                                   \\ \cline{2-3}
                                                                              & Fixed              & Moving            &                                &                                                                                                       \\ \cline{2-3}
                                                                              & Moving             & Moving            &                                &                                                                                                       \\ \hline
\end{tabular}
\end{table}

The node and edge features, along with their detailed interpretations, are presented in Tab. \ref{table:node_feat} and Tab. \ref{table:edge_feat}, respectively.

The raw node features are first passed through a projection layer, mapping nodes with different feature dimensions into a common high-dimensional space.

\begin{equation}
\begin{aligned}
\mathbf{n}_{i} = \mathit{f}_{pro}(^{raw}\mathbf{n}_{i})
\end{aligned}
\end{equation}
where $^{raw}\mathbf{n}_{i}$ and $\mathbf{n}_{i}$ refer to the raw and projected node features of node $i$ respectively.
The EGAT layer then performs an attentional aggregation, where the attention coefficients and the aggregated features are computed by jointly considering the self-node, neighboring nodes, and the connecting edge features. 

\begin{equation}
\begin{aligned}
\mathbf{n}_{i}^{\prime} = & \boldsymbol{\Theta}_{\mathrm{self}} \cdot \mathbf{n}_{i}+\sum_{j \in \mathcal{N}\left(n_{i}\right)} \alpha_{j, i}\left(\boldsymbol{\Theta}_{\mathrm{neigh}} \cdot \mathbf{n}_{j}+\boldsymbol{\Theta}_{\mathrm{edge}} \cdot \mathbf{e}_{j, i}\right) \\
\alpha_{j, i} = & soft(Leak(\boldsymbol{a}^T[\boldsymbol{\Theta}_{\mathrm{self}} \cdot \mathbf{n}_{i} || \boldsymbol{\Theta}_{\mathrm{neigh}} \cdot \mathbf{n}_{j} || \boldsymbol{\Theta}_{\mathrm{edge}} \cdot \mathbf{e}_{j, i}]))
\end{aligned}
\end{equation}
where $\boldsymbol{\Theta}_{\mathrm{self}}$, $\boldsymbol{\Theta}_{\mathrm{neigh}}$ and $\boldsymbol{\Theta}_{\mathrm{edge}}$ denote the learnable weight matrices for the feature transformation of self-node ($\mathbf{n}_{i}$), neighbor node ($\mathbf{n}_{j}$) and connected edge ($\mathbf{e}_{j, i}$). $\boldsymbol{a}^T$ is a learnable vector to transform the aggregated features and yield attention weight $\alpha_{j, i}$. The $soft$ and $Leak$ refer to the standard softmax and LeakyReLU processes.

\subsection{Details of Multi-strategy Multi-frame Sinkhorn Assignment}
\label{subsec:app-sinkhorn}

\subsubsection{Split} 
For frame $k$, the prior-camera score matrix $^{k}\mathbf S\in \mathbb{R}^{N_R\times N_C}$ is calculated by the dot product between their post-GNN aggregated embeddings.
\begin{equation}
    \begin{aligned}
    ^{k}\textbf{S} _{i,j} = <^{k}\textbf{f}^{\mathcal{P}}_{i}, ^{k}\textbf{f}^{\mathcal{D}}_{j}>
    \end{aligned}
\end{equation}
where $^{k}\textbf{S} _{i,j}$ indicated the element at entry $(i,j)$ of the score matrix for the $k$-th frame, $^{k}\textbf{f}^{\mathcal{P}}_{i}$, $^{k}\textbf{f}^{\mathcal{D}}_{j}$ denote the $i$-th prior bearing features and the $j$-th camera bearing features associated with the $k$-th frame.
An augmentation is applied to the score matrix by adding a dedicated \textit{bin} row and column to accommodate discarded observations, producing the augmented score matrix $^{k}\bar{\mathbf S}\in \mathbb{R}^{(N_R+1)\times (N_C+1)}$.
The augmented score matrix is subsequently iteratively normalized by the Sinkhorn algorithm, which redistributes the scores under the row and column sum constraints, forming an optimal local assignment matrix $_{(N_I)}^{k}\bar{\mathbf S}$, where $N_I$ denotes the iteration numbers.
Finally, we select the bearing candidate with the highest score from each row and reject incorrect observations using a predefined threshold.

\begin{figure}[tp]
    \centering
    \includegraphics[width=.48\textwidth]{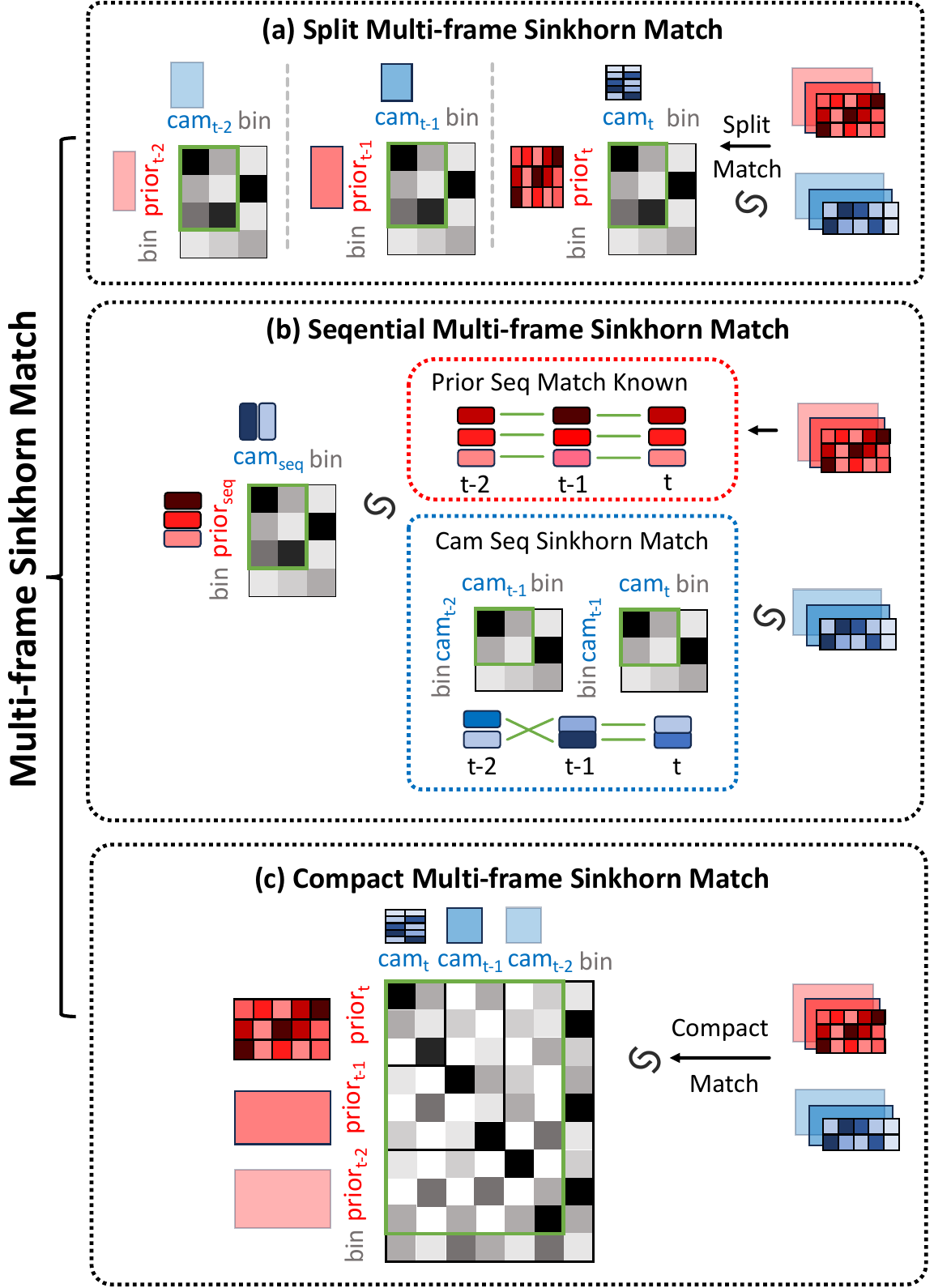}
    \caption{Multi-strategy Sinkhorn assignment. The figure demonstrates the matching procedure for a scenario involving three robots, two camera observations, and a time window of three frames.}
    \label{fig:matchnet}
\end{figure}

\subsubsection{Sequential} 
The sequential matching problem comprises two stages: the first stage performs intra-set matching within the prior and camera sets, respectively, while the second stage conducts cross-sequence matching between the resulting prior and camera sequences.
During the first-stage matching, since the prior bearings have known identities across frames, only inter-frame matching among anonymous multi-frame camera bearings is required. After the intra-set inter-frame correspondences of the prior and camera bearings are determined, a shared aggregation layer fuses information across frames to construct sequence-level representations, followed by a seq-to-seq matching between the prior and camera sequences.

\subsubsection{Compact} 
It compacts bearing features from all frames into a single score matrix $^{k}\mathbf S\in \mathbb{R}^{(N_R*N_W)\times (N_C*N_W)}$.
Compared with the frame-to-frame matching strategy in $\textit{Split}$, the $\textit{Compact}$ method makes more effective use of temporal information during partial assignment. In contrast to the seq-to-seq matching in $\textit{Sequential}$, it eliminates the redundant intra-set sequence matching process and avoids the failure of cross-sequence matching caused by potentially inaccurate camera sequence features, achieving unified matching in a single step.

\begin{table*}[htbp]
\caption{Four node types in heterogeneous EGAT.}
\label{table:node_feat}
\renewcommand\arraystretch{1.6}
\centering
\begin{tabular}{cc|cc}
\hline
\multirow{2}{*}{\textbf{Node Type}} & \multirow{2}{*}{\textbf{Feat}} & \multicolumn{2}{c}{\textbf{Description}}                                                                                               \\ \cline{3-4} 
                                    &                                    & \multicolumn{1}{c|}{\textbf{Single-robot}}                        & \textbf{Multi-robot}                                               \\ \hline
Ref                                 & $\mathbf P_{B_r}^{B_r}$                                  & \multicolumn{1}{c|}{World frame}                                  & Base frame of the reference robot $r$                                  \\ \hline
Fixed                               & $\mathbf t_{T_{r,n}}^{B_r}$                                   & \multicolumn{1}{c|}{Anchor $n$'s position in the world frame}           & Tag $n$'s extrinsic of the reference robot $r$                               \\ \hline
Base                                & $\mathbf {\hat P}_{B_i}^{B_r}$                                  & \multicolumn{1}{c|}{Base pose of the robot in the world frame}    & Base poses of the other estimated robots $i$ in the reference frame    \\ \hline
Moving                              & $\mathbf {\hat t}_{T_{i,m}}^{B_r}$                                  & \multicolumn{1}{c|}{Tag $m$'s position of the robot in the world frame} & Tag $m$'s positions of the other estimated robots $i$ in the reference frame \\ \hline
\end{tabular}
\end{table*}

\subsection{Datasets and Hyperparameter Settings}
\label{subsec:app-data}

\begin{table*}[!ht]
\caption{Multi-scenario multi-task simulation and real-world datasets for pre-training and testing. Each dataset category may include multiple scenes, with each scene comprising one or more trajectory sequences. Although the test dataset may correspond to the seen dataset category used during pre-training, all test scenes for fine-tuning are strictly unseen in the pre-training set.}
\label{table:datasets}
\renewcommand\arraystretch{1.2}
\setlength{\tabcolsep}{3.0pt}
\centering
\begin{tabular}{cc|cccccc|ccccccc}
\hline
\multicolumn{2}{c|}{\textbf{Scene}}                                                                                                          & \textbf{Pretrain}         & \textbf{Test}             & \textbf{Sim}              & \textbf{\begin{tabular}[c]{@{}c@{}}With \\ Bearing\end{tabular}} & \textbf{\begin{tabular}[c]{@{}c@{}}Spatial\\ Anchors\end{tabular}} & \textbf{NLOS}             & \textbf{\begin{tabular}[c]{@{}c@{}}Scene\\ Num\end{tabular}} & \textbf{\begin{tabular}[c]{@{}c@{}}Seq\\ Num\end{tabular}} & \textbf{\begin{tabular}[c]{@{}c@{}}Robot\\ Num\end{tabular}} & \textbf{\begin{tabular}[c]{@{}c@{}}Tag\\ Num\end{tabular}} & \textbf{\begin{tabular}[c]{@{}c@{}}Anchor\\ Num\end{tabular}} & \textbf{\begin{tabular}[c]{@{}c@{}}Sample\\ Num\end{tabular}} & \textbf{Area}                \\ \hline
\multirow{7}{*}{\textbf{\begin{tabular}[c]{@{}c@{}}Single\\ Robot\end{tabular}}} & \begin{tabular}[c]{@{}c@{}}circle1\end{tabular} & \checkmark & \checkmark & \checkmark & \ding{55}                                           & \checkmark                                          & \ding{55}    & 24                                                           & 98                                                         & 1                                                            & 1$\sim$4                                                   & 3$\sim$8                                                      & 91314                                                         & $20m \times 20m \times 2m$                                                            \\ \cline{2-15} 
                                                                                 & \begin{tabular}[c]{@{}c@{}}circle2\end{tabular}  & \checkmark & \ding{55}    & \checkmark & \ding{55}                                           & \ding{55}                                             & \ding{55}    & 25                                                           & 100                                                        & 1                                                            & 1$\sim$4                                                   & 3$\sim$8                                                      & 94900                                                         & $70m \times 12m \times 2m$                                                           \\ \cline{2-15} 
                                                                                 & square\cite{kao2023viunet}                                                    & \ding{55}    & \checkmark & \ding{55}    & \ding{55}                                           & \ding{55}                                             & \ding{55}    & 1                                                            & 2                                                          & 1                                                            & 1                                                          & 4                                                             & 257                                                           & $5m \times 2m \times 1m$                                                            \\ \cline{2-15} 
                                                                                 & corridor\cite{wang2025neural}                                                  & \checkmark & \ding{55}    & \ding{55}    & \ding{55}                                           & \ding{55}                                             & \ding{55}    & 1                                                            & 1                                                          & 1                                                            & 4                                                          & 4                                                             & 23                                                            & $50m \times 4m \times 1m$                                                            \\ \cline{2-15} 
                                                                                 & outdoor\cite{wang2025neural}                                                     & \checkmark & \ding{55}    & \ding{55}    & \ding{55}                                           & \ding{55}                                             & \ding{55}    & 1                                                            & 1                                                          & 1                                                            & 4                                                          & 4                                                             & 158                                                           & $40m \times 4m \times 3m$                                                             \\ \cline{2-15} 
                                                                                 & tunnel1\cite{wang2025neural}                                                    & \checkmark & \ding{55} & \ding{55}    & \ding{55}                                           & \ding{55}                                             & \ding{55}    & 1                                                            & 11                                                         & 1                                                            & 4                                                          & 36                                                            & 3367                                                          & $330m \times 60m \times 3m$                              \\ \cline{2-15}              
                                                                                 & tunnel2                                 & \ding{55}     & \checkmark    & \ding{55}   & \ding{55}   & \ding{55}  & \ding{55}    & 1    & 4   & 1  & 4  & 13   & 1266   & $120m \times 6m \times 3m$  \\ \hline
\multirow{5}{*}{\textbf{\begin{tabular}[c]{@{}c@{}}Multi\\ Robot\end{tabular}}}  & spiral                                                    & \checkmark & \checkmark & \checkmark & \checkmark                                        & -                                                                  & \ding{55}    & 18                                                           & 56                                                         & 3$\sim$8                                                     & 1$\sim$4                                                   & 0                                                             & 13682                                                         & $20m \times 20m \times 2m$                                                            \\ \cline{2-15} 
                                                                                 & \begin{tabular}[c]{@{}c@{}}random\\ forest\end{tabular}   & \checkmark & \ding{55}    & \checkmark & \checkmark                                        & -                                                                  & \checkmark & 38                                                           & 38                                                         & 4$\sim$16                                                    & 1                                                          & 0                                                             & 3187                                                          & $50m \times 10m \times 2m$                                                            \\ \cline{2-15} 
                                                                                 & lab-los\cite{wang2025mr}                                                   & \ding{55}    & \checkmark & \ding{55}    & \checkmark                                        & -                                                                  & \ding{55}    & 1                                                            & 2                                                          & 5                                                            & 1                                                          & 0                                                             & 858                                                           & $15m \times 4m \times 1m$                                                             \\ \cline{2-15} 
                                                                                 & lab-nlos\cite{wang2025mr}                                                  & \ding{55}    & \checkmark & \ding{55}    & \checkmark                                        & -                                                                  & \checkmark & 1                                                            & 2                                                          & 5                                                            & 1                                                          & 0                                                             & 542                                                           & $15m \times 4m \times 1m$                                                             \\ \cline{2-15} 
                                                                                 & plane\cite{fishberg2024murp}                                                      & \ding{55}    & \checkmark & \ding{55}    & \ding{55}                                           & -                                                                  & \ding{55}    & 1                                                            & 2                                                          & 3                                                            & 6                                                          & 0                                                             & 1035                                                          & $10m \times 10m \times 2m$                                                            \\ \hline
\multicolumn{2}{c|}{\textbf{TOTAL}}                                                                                                          & \multicolumn{6}{c|}{Both}                                                                                                                                                                                                                             & 113                                                          & 317                                                        & 1$\sim$16                                                    & 1$\sim$6                                                   & 0$\sim$36                                                     & 210589                                                        & -                                                                                     \\ \hline
\end{tabular}
\end{table*}

The multi-scenario dataset is presented in Tab. \ref{table:datasets} and the details of our test dataset are illustrated in Tab. \ref{table:testdata}, which are constructed using Deep Graph Library (DGL) \cite{wang2019dgl}, adhering to the unified heterogeneous graph architecture we introduced in Section \ref{subsec:egat}. Our datasets cover a variety of experimental setups (handheld, UAV, UGV carriers; indoor/outdoor, occluded/non-occluded conditions). With our unified formulation, data from diverse training scenarios are transformed into a common data type and then batched and shuffled randomly during training, ensuring diverse and unbiased learning, without requiring any specialized process for scene transfer.

We use 225 trajectories for training and 76 for validation. For evaluation across the 4 different localization tasks, we construct a test set of 8 scenes (2 scenes per task), including 6 real-world unseen scenes, ensuring that each task includes at least one real-world unseen scene. Each scene comprises two trajectories (one for fine-tuning, one for testing).

\begin{table}[tp]
\caption{Test dataset details.}
\label{table:testdata}
\setlength{\tabcolsep}{3.5pt}
\centering
\begin{tabular}{c|cccccccc}
\hline
                    \begin{tabular}[c]{@{}c@{}}Test \\ Dataset\end{tabular} & \begin{tabular}[c]{@{}c@{}}sim-\\ circle\end{tabular} & square & \begin{tabular}[c]{@{}c@{}}tunnel\\ drive\end{tabular} & \begin{tabular}[c]{@{}c@{}}tunnel\\ fly\end{tabular} & \begin{tabular}[c]{@{}c@{}}lab-\\ los\end{tabular} & \begin{tabular}[c]{@{}c@{}}lab-\\ nlos\end{tabular} & \begin{tabular}[c]{@{}c@{}}sim-\\ spiral\end{tabular} & plane \\ \hline
\textbf{Real World}  & \ding{55}                                                    & \checkmark    & \checkmark                                                    & \checkmark                                                  & \checkmark                                                & \checkmark                                                 & \ding{55}                                                    & \checkmark   \\
\textbf{Unseen} & \ding{55}                                                    & \checkmark    & \checkmark                                                    & \checkmark                                                  & \checkmark                                                & \checkmark                                                 & \ding{55}                                                    & \checkmark   \\ \hline
\end{tabular}
\end{table}

In the EGAT network, the number of attention heads is set to 3, the number of GNN layers to 4, and the hidden layer dimension to 64. The GRU uses 50 frames of UWB range data for filtering.
We use the Adam optimizer, with a learning rate of 1e-3, a weight decay coefficient of 5e-4, and a batch size of 4096 for pretraining. All pretraining and fine-tuning are conducted on a single RTX 3090 GPU.

For hyperparameters in the loss functions, we adopt a simple ($\lambda_p$ = $\lambda_r$ = $\lambda_q$ = 1.0) or principled ($\lambda_m$ = 10.0 and $\lambda_d$ = 0.04) setting. Note that both $\lambda_m$ and $\lambda_d$ are directly inherited from the public implementation of Mr. Virgil \cite{wang2025mr}, and we observe stable generalization without further tuning.
Regarding the parameter $\lambda_d$, which appears in nearly all loss functions, we conducted a hyperparameter sensitivity test with four configurations (0.01, 0.04, 1.0, 10.0), training on the pretraining dataset and evaluating the RMSE error across 108 scenes. The error variation is within 3 cm, indicating that the model is robust to the hyperparameter.

\subsection{Baselines}
\label{subsec:app-baseline}

We include the generic PGO solver as a universal baseline and further choose one task-specific comparison method for each individual localization setting:
\begin{itemize}
  \item 
    $\textit{VIUNet}$ \cite{kao2023viunet} combines CNN and LSTM modules to fuse VIO and UWB data for single-robot, single-tag localization. To maintain consistent input information among all compared methods, we compare the UWB branch.
  \item 
    $\textit{NRIO}$ \cite{wang2025neural} consists of a GNN-based UWB localization network and a GRU-based neural inertial odometry module for single-robot multi-tag localization. In our evaluation, we benchmark the GNN UWB localization network.
  \item 
    $\textit{Mr. Virgil}$ \cite{wang2025mr} is an end-to-end multi-robot, single-tag localization network based on UWB ranging and anonymous visual detections. Note that it incorporates a pseudo-visual odometry (PVO) module to assist localization, potentially affecting deployment in practical systems. In our experimental setup, we exclude the PVO assistance.
  \item 
    $\textit{MURP}$ \cite{fishberg2024murp} performs multi-robot, multi-tag localization using local UWB measurements and prior state constraints in a least-squares framework. To maintain experimental consistency, we evaluate the variant that incorporates learnable UWB bias correction and Huber loss optimization, but does not enforce a fixed z-axis.
  \item 
    $\textit{PGO}$, as a general least-squares-based pose graph optimization method, can be widely applied to various localization tasks.
\end{itemize}

\begin{figure*}[htbp]
    \centering
    \includegraphics[scale=0.38]{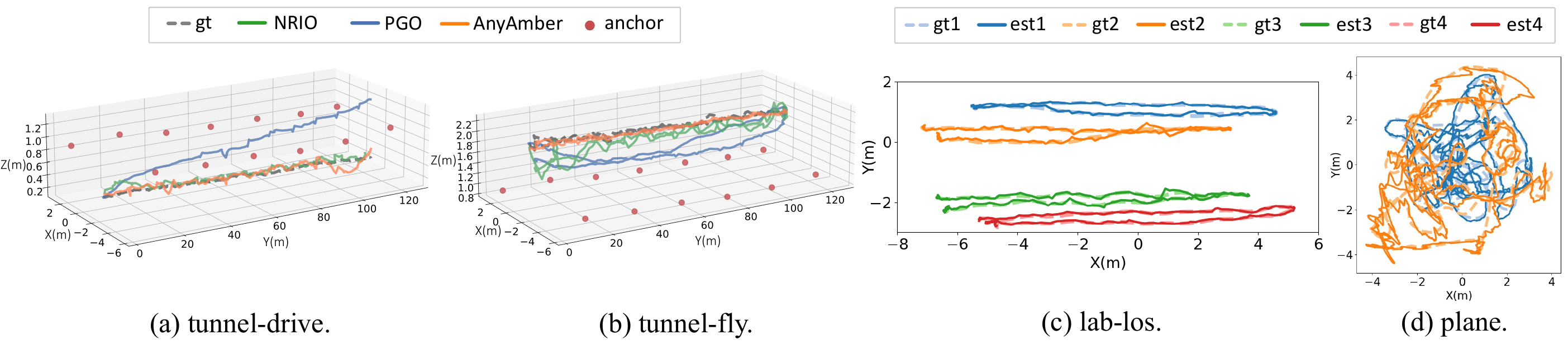}
    \caption{Visualization of additional trajectory estimates in single-robot (a–b) and multi-robot (c–d) scenarios.}
    \label{fig:traj-part2}
\end{figure*}

\subsection{Additional Localization Experiments}
\label{subsec:addi-exp}

Fig. \ref{fig:traj-part2} presents additional trajectory comparisons from real-world localization experiments. We also expand our evaluation to include the public real-world RCV dataset \cite{choe2024addressing}, which features 6 varied scenes, each with different anchor layouts, trajectories, and noise characteristics. 
Following the same data split as VIUNet, we fine-tune our model on one subset and test on the other, demonstrating notably higher accuracy over the baselines. Tab. \ref{table:rcv} shows the positioning error.

\begin{table}[tp]
\caption{RMSE (m) across all 6 sequences of RCV dataset.}
\label{table:rcv}
\centering
\begin{tabular}{c|cccccc|c}
\hline
                  & Seq1          & Seq2          & Seq3          & Seq4          & Seq5          & Seq6          & Avg.       \\ \hline
PGO               & 2.37          & 5.28          & 1.77          & 1.50          & 0.43          & 4.02          & 2.56          \\
VIUNet            & 0.91          & 3.54          & 0.71          & 0.27          & 0.18          & 0.37          & 1.00          \\
\textbf{AnyAmber} & \textbf{0.37} & \textbf{0.17} & \textbf{0.15} & \textbf{0.22} & \textbf{0.12} & \textbf{0.27} & \textbf{0.22} \\ \hline
\end{tabular}
\end{table}

\begin{figure}[tp]
    \centering
    \subfigure[UWB ranging variation.]{
    \includegraphics[width=.45\textwidth]{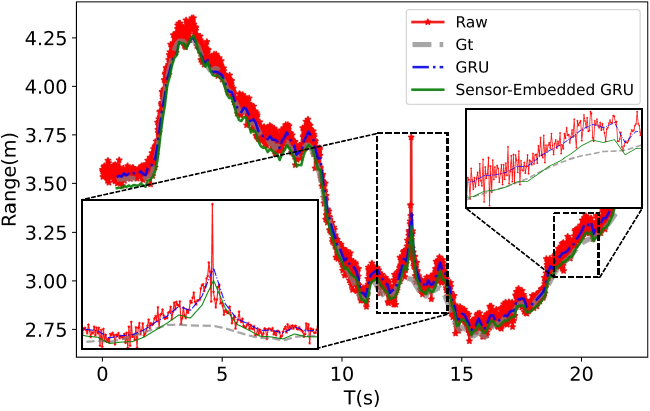}
    }
    \subfigure[Ranging error comparison.]{
    \includegraphics[width=.45\textwidth]{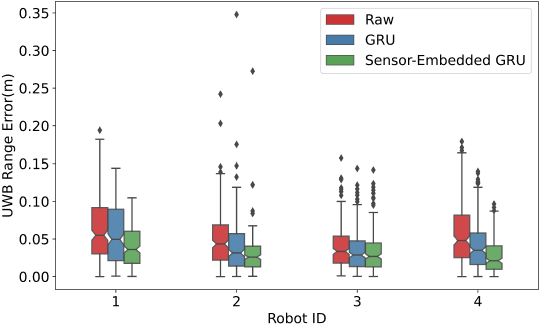}
    }
    \caption{Sensor-Embedded GRU range filter and ranging error comparison in real-world multi-robot environments.}
    \label{fig:gru-r5t1-los6}
\end{figure}

\vspace{-0.5em}

\subsection{Functional Validation of Network Components}
\label{subsec:app-func-valid}

\subsubsection{Sensor-Embedded GRU Range Filter} 
A GRU model without UWB sensor embeddings is pre-trained on scenarios with varying noise and bias, including both simulated and real-world data. During fine-tuning, the sensor embedding is introduced, enabling the model to learn the specific noise characteristics of the target scenario, thereby improving filtering and variance estimation accuracy. 

Fig. \ref{fig:gru-r5t1-los6} demonstrates the filtering performance and error correction in a real-world multi-robot, single-tag scenario. Owing to the hand-held operation for indoor data collection, the UWB ranging measurements were subjected to body occlusion, multi-path effects, and antenna deployment issues, resulting in noticeable noise and bias.
A vanilla GRU can effectively smooth the data and mitigate abrupt outliers, but it cannot eliminate range biases that stem from sensor-dependent characteristics. By incorporating UWB sensor embeddings, the Sensor-Embedded GRU is able to jointly perform filtering and bias compensation, achieving superior ranging calibration.

\begin{table*}[!ht]
\caption{Multi-robot matching results in both simulated and real-world scenarios with different numbers of robots (indicated in parentheses). Methods shown in bold denote the proposed temporal graph neural network and the suite of multi-frame matching algorithms. The suffix after \textit{win} denotes the number of frames in the matching window. The best F1 score is highlighted in bold, while the second-best is marked with underlining. }
\label{table:match}
\renewcommand\arraystretch{1.2}
\setlength{\tabcolsep}{8pt}
\centering
\begin{tabular}{ccccccccccccc}
\hline
\multirow{3}{*}{Method} & \multicolumn{12}{c}{Datasets}                                                                                                                                                                      \\ \cline{2-13} 
                        & \multicolumn{3}{c|}{Sim-Forest(8)}                  & \multicolumn{3}{c|}{Sim-Forest(16)}                 & \multicolumn{3}{c|}{Real-LOS(5)}                    & \multicolumn{3}{c}{Real-NLOS(5)} \\ \cline{2-13} 
                        & P\%   & R\%   & \multicolumn{1}{c|}{F1\%}           & P\%   & R\%   & \multicolumn{1}{c|}{F1\%}           & P\%   & R\%   & \multicolumn{1}{c|}{F1\%}           & P\%    & R\%    & F1\%           \\ \hline
handcraft @ 0.9         & 76.48 & 99.55 & \multicolumn{1}{c|}{86.50}          & 55.89 & 98.99 & \multicolumn{1}{c|}{71.44}          & 91.30 & 99.70 & \multicolumn{1}{c|}{95.32}          & 87.97  & 99.89  & 93.55          \\
handcraft @ 0.99        & 95.32 & 98.63 & \multicolumn{1}{c|}{96.95}          & 89.85 & 98.04 & \multicolumn{1}{c|}{93.77}          & 99.64 & 92.26 & \multicolumn{1}{c|}{95.81}          & 99.70  & 95.85  & 97.74          \\ \hline
GNN-win1                & 98.37 & 98.61 & \multicolumn{1}{c|}{98.49}          & 95.39 & 95.43 & \multicolumn{1}{c|}{95.41}          & 98.57 & 99.15 & \multicolumn{1}{c|}{98.86}          & 98.69  & 99.13  & 98.91          \\ \hline
TGNN-split-win3         & 99.36 & 98.88 & \multicolumn{1}{c|}{99.12}          & 96.42 & 95.61 & \multicolumn{1}{c|}{96.01}          & 98.93 & 99.51 & \multicolumn{1}{c|}{99.22}          & 98.94  & 99.59  & 99.26          \\
TGNN-split-win5         & 99.23 & 99.07 & \multicolumn{1}{c|}{99.15}          & 97.38 & 96.32 & \multicolumn{1}{c|}{96.85}          & 99.24 & 99.44 & \multicolumn{1}{c|}{99.34}          & 99.38  & 99.34  & 99.36          \\ \hline
TGNN-seq-win3           & 99.48 & 99.37 & \multicolumn{1}{c|}{99.42}          & 98.33 & 98.23 & \multicolumn{1}{c|}{{\underline{98.28}}}    & 99.35 & 99.56 & \multicolumn{1}{c|}{\textbf{99.45}} & 99.39  & 99.78  & {\underline{99.58}}    \\
TGNN-seq-win5           & 99.45 & 99.51 & \multicolumn{1}{c|}{{\underline{99.48}}}    & 98.49 & 98.06 & \multicolumn{1}{c|}{98.27}          & 99.31 & 99.57 & \multicolumn{1}{c|}{{\underline{99.44}}}    & 99.40  & 99.82  & \textbf{99.61} \\ \hline
TGNN-compact-win3       & 99.49 & 99.67 & \multicolumn{1}{c|}{\textbf{99.58}} & 98.57 & 98.64 & \multicolumn{1}{c|}{\textbf{98.60}} & 99.07 & 99.48 & \multicolumn{1}{c|}{99.27}          & 99.09  & 99.65  & 99.37          \\
TGNN-compact-win5       & 99.42 & 99.39 & \multicolumn{1}{c|}{99.40}          & 98.56 & 97.97 & \multicolumn{1}{c|}{98.26}          & 99.02 & 99.33 & \multicolumn{1}{c|}{99.17}          & 99.02  & 99.59  & 99.30          \\ \hline
\end{tabular}
\end{table*}

\subsubsection{Multi-robot Multi-frame Data Association}
We evaluate different multi-robot matching methods through precision, recall rate and F1 scores under diverse conditions: in simulation and real-world, with and without occlusion, and across different robot counts.
The baselines considered in our evaluation consist of the handcrafted nearest-neighbor–based matching methods with different outlier rejection thresholds (\textit{handcraft @ 0.9} and \textit{handcraft @ 0.99}), along with the single-frame GNN matching approach introduced in Mr. Vrigil\cite{wang2025mr} (\textit{GNN-win1}).
Specifically, \textit{handcraft @ 0.9} means that only matches with a bearing cosine similarity above 0.9 are retained; the interpretation for \textit{handcraft @ 0.99} follows accordingly. Tab. \ref{table:match} summarizes the comparative performance of various matching algorithms across the multi-scenario dataset.

\textbf{Temporal feature aggregation:}
Both \textit{TGNN-split-winX} and \textit{GNN-win1} adopt the same frame-to-frame matching paradigm in the Sinkhorn matching stage.
The key difference lies in the feature aggregation front-end: the former incorporates temporal information into feature aggregation. The superior F1 score of TGNN-based methods demonstrates that incorporating temporal context significantly strengthens the representational capacity of the graph-based feature aggregation.

\textbf{Multi-frame matching:} 
 Compared with \textit{TGNN-split-winX}, which uses a single-frame matching manner, \textit{TGNN-seq-winX} and \textit{TGNN-compact-winX} exploit multi-frame feature information during the assignment process. As a result, despite using the same TGNN aggregation front-ends, these two variants exhibit superior matching results. Notably, the \textit{TGNN-compact-winX} approach eliminates the need for additional camera bearing matching and sequential feature aggregation, enabling multi-frame information to be jointly optimized within a single score matrix. To balance matching performance and complexity, we finally adopt the \textit{TGNN-compact-win3} method throughout the localization pipeline.
 
\begin{figure}[tp]
    \centering
    \includegraphics[width=.48\textwidth]{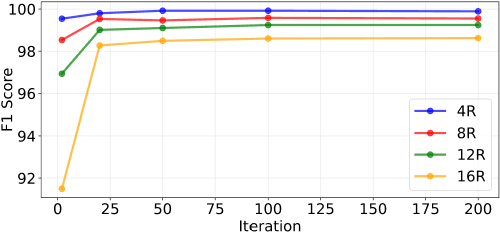}
    \caption{Matching F1 score versus Sinkhorn iteration number in scenarios with different robot scales.}
    \label{fig:f1}
\end{figure}

\textbf{Sinkhorn iterations:}
 We assess the effect of Sinkhorn iteration number on matching outcomes by applying the \textit{TGNN-compact-win3} approach with different iteration settings to datasets comprising 4 to 16 robots. The corresponding results are presented in Fig. \ref{fig:f1}. To achieve a trade-off between matching accuracy and execution efficiency, 100 Sinkhorn iterations are employed in our final setup.

\subsection{Ablation Study on Sensor Input}

To examine how ranging and bearing measurements influence localization performance, we evaluate the effect of different sensor inputs on localization accuracy in a single-robot simulation scenario with four anchors. We guarantee the same test trajectory and similar prior poses generated by adding noise with a standard deviation of 0.4 m and 0.04 rad to the ground truth. Tab. \ref{table:ablation-sensor} demonstrates that incorporating multi-tag ranging measurements and additional robot-to-anchor bearing observations consistently enhances the localization accuracy of both EGAT and EGAT+RPGO methods.

\begin{table}[tp]
\caption{Ablation study results on sensor inputs.}
\label{table:ablation-sensor}
\renewcommand\arraystretch{1.2}
\setlength{\tabcolsep}{8pt}
\centering
\begin{tabular}{cccc}
\hline
\textbf{Setting}   & \textbf{Single-Tag} & \textbf{Multi-Tag} & \textbf{\begin{tabular}[c]{@{}c@{}}Multi-Tag\\ w. bearing\end{tabular}} \\ \hline
EGAT      & 0.143               & 0.111              & 0.087                                                                   \\ \hline
EGAT+RPGO & 0.140               & 0.100              & \textbf{0.083}                                                          \\ \hline
\end{tabular}
\end{table}

\end{document}